%% file: bare_conf.tex
\documentclass[conference]{IEEEtran}
\ifCLASSINFOpdf
   \usepackage[pdftex]{graphicx}
   \usepackage{subfig}
   \usepackage{amsfonts}
   \usepackage{url}
	\usepackage{booktabs}
\else
\fi
\usepackage{amsmath}
\usepackage{commath}
\makeatletter
\def\ps@IEEEtitlepagestyle{%
	\def\@oddfoot{\mycopyrightnotice}%
	\def\@evenfoot{}%
}
\def\mycopyrightnotice{%
	{\footnotesize This work has been submitted to the IEEE for possible publication. Copyright may be transferred without notice, after which this version may no longer be accessible.\hfill}
	\gdef\mycopyrightnotice{}
}

\usepackage{booktabs}

\usepackage{hyperref}

\begin{document}
%
\title{Lightweight Deep Learning Architecture for MPI Correction and Transient Reconstruction}



%
\author{\IEEEauthorblockN{Adriano Simonetto\IEEEauthorrefmark{1},
Gianluca Agresti\IEEEauthorrefmark{2},
Pietro Zanuttigh\IEEEauthorrefmark{1} and
Henrik Schäfer\IEEEauthorrefmark{2}}
\IEEEauthorblockA{\IEEEauthorrefmark{1}Department of Information Engineering,
University of Padova, Padova, Italy,
\\Email: adriano.simonetto@phd.unipd.it, zanuttigh@dei.unipd.it}
\IEEEauthorblockA{\IEEEauthorrefmark{2} R$\&$D Center Europe Stuttgart Laboratory 1, Stuttgart, Germany,
\\Email: Gianluca.Agresti@sony.com, Henrik.Schaefer@sony.com}}


\maketitle

\begin{abstract}
Indirect Time-of-Flight cameras (iToF) are low-cost devices that provide depth images at an interactive frame rate. However, they are affected by different error sources, with the spotlight taken by Multi-Path Interference (MPI), a key challenge for this technology. Common data-driven approaches tend to focus on a direct estimation of the output depth values, ignoring the underlying transient propagation of the light in the scene. In this work instead, we propose a very compact architecture, leveraging on the direct-global subdivision of transient information for the removal of MPI and for the reconstruction of the transient information itself. The proposed model reaches state-of-the-art MPI correction performances both on synthetic and real data and proves to be very competitive also at extreme levels of noise; at the same time, it also makes a step towards reconstructing transient information from multi-frequency iToF data.
\end{abstract}



%
\IEEEpeerreviewmaketitle

\input{sections/introduction}
\input{sections/related}

\input{sections/model}
\input{sections/method}

\input{sections/datasets}

\input{sections/results}

\input{sections/conclusions}

\bibliographystyle{IEEEtran}
\bibliography{IEEEabrv,bare_conf}

\end{document}

%% file: sections/introduction.tex
\section{Introduction}
\label{sec:introduction}

The demand for more accurate and reliable range imaging devices has seen a constant rise over the years. Their applications are widespread ranging from autonomous driving \cite{auto1,auto2} to augmented reality \cite{augmented}, 3D reconstruction \cite{rec1,rec2} and even landing on planetary bodies \cite{moon}. 
The working principles are different for the various types of sensors, but the main objective remains the same: retrieving the distance information between the camera and the target object. Some of the most common technologies are stereo imaging \cite{stereo}, where the depth information is retrieved from a couple of RGB cameras at a fixed distance, Time-of-Flight (ToF) based devices \cite{tof_survey}, e.g. LIDARs \cite{lidar} or matrix ToF sensors, and structured light scanners \cite{struct}, that rely on light patterns.
In this work we will focus our attention on ToF based technologies, more specifically on indirect Time-of-Flight (iToF) cameras. A direct Time-of-Flight (dToF) device sends an impulse of light towards the scene, measures the travel time of the impulse and computes the depth information from that. An iToF camera instead sends a modulated light signal and correlates the reflected signal with the sensor modulation signal; from these measures the distance is retrieved.
iToF-based cameras are quite accurate, have a good spatial resolution and are nowadays sold at consumer level in some of the most recent mobile phones \cite{xperia}. This technology however also has a significant drawback, which is intrinsic to its basic operating principle and is called Multi-Path Interference (MPI). This error source typically produces an overestimation of the depth, which is strictly linked to the scene geometry and that has been widely studied in the literature \cite{sra,su,ag_2018,buratto}. Some early single frequency approaches such as \cite{single1,single2} proposed an algorithmic solution for the problem, but also had to set some unrealistic assumptions (e.g. knowing the scene structure) in order to make it tractable. Following attempts highlighted the need for multiple iToF acquisitions at different frequencies in order to better deal with MPI \cite{multi1,multi2}, and also linked the iToF and dToF domains showing that it is possible to go from dToF to iToF with a simple linear model \cite{sra}. 
The true turning point however, came when deep learning started being applied to the field. The first deep learning architectures \cite{deeptof,su} improved on the previous works but were still quite complex models that at the same time did not perform well on real data. The main issue is that the acquisition of real iToF data with matching depth ground truth is a challenging task, and synthetic images are therefore the main training tool. This problem has been tackled in \cite{ag_unsup} where an Unsupervised Domain Adaptation (UDA) approach was proposed showing that it is possible to improve the model generalization without the need for real ground truth; in \cite{ag_pami} they expanded the method by considering different domain adaptation scenarios. Recently, a couple of data-driven approaches exploiting the relationship between iToF and dToF information \cite{buratto,itof2dtof} have been proposed.

In this work, we introduce a novel modular deep learning approach that leverages on the dual nature of transient information for MPI correction and for the estimation of dToF data. The model is composed of three parts, the first one needed for dealing with temporal noise sources with zero mean such as shot noise, the second built for MPI denoising and the final module instead for the reconstruction of the transient representation. We propose in particular two architectures. The first, \textit{SD}, reaches state-of-the-art performance both on synthetic and real data, is able to deal with extreme amounts of zero-mean noise and is much lighter than the best performing networks in the literature; the second, \textit{D}, instead falls only a little behind in performance on real data, but is extremely lightweight, i.e. it has only $3k$ parameters. An additional contribution of this paper is the introduction of the \textit{Walls} dataset; it is a novel transient dataset based on simple geometries that is used for training both the modules for MPI denoising and the one for transient reconstruction. 

The rest of this paper is structured as follows: Section \ref{sec:related} gives an overview of the current works on MPI denoising and transient reconstruction; Section \ref{sec:model} explains the working principle of iToF cameras and introduces the notation, while Section \ref{sec:method} instead shows the proposed architecture and describes its modules; in Section \ref{sec:datasets} we discuss the employed datasets, focusing especially on the one that we introduce; finally, in Section \ref{sec:results} we give an in depth quantitative and qualitative comparison with some state-of-the-art approaches, and in Section \ref{sec:conclusions} we draw our conclusions and describe some future developments.

%% file: sections/related.tex
\section{Related Works}
\label{sec:related}


As remarked in Section \ref{sec:introduction}, Multi-Path Interference is a non-zero mean error source that is intrinsic to the iToF technology, and at the same time one of its key limitations. The approaches that tackle MPI correction can be generally divided into two groups, single-frequency and multi-frequency ones. Those belonging to the first group such as \cite{single1,single2} and \cite{single3} exploit a reflection model together with the spatial information provided by the MPI-corrupted image for their solution. Jimenez et al \cite{single3} for example, proposed an iterative optimization algorithm based on the assumption that all scene surfaces are perfectly Lambertian. 
Early multi-frequency approaches show similar constraints. In \cite{sra}, Freedman et al. introduced the relationship between iToF measurements and the transient behaviour of light extending the problem to the case of $K$ interfering rays. They then proposed an algorithm for MPI correction treating it as an $L_1$ optimization problem. Bhandari et al. \cite{multi2}, adopted similar assumptions but offered instead a non-iterative solution using Vandermonde matrices.

The restrictions of these models and the unrealistic amount of input frequencies required for the solutions lead to a rapid rise in popularity of deep learning based approaches. Marco et al. \cite{deeptof} proposed an encoder-decoder architecture with a split training approach: the encoder was trained on unlabelled real data, and the decoder on the synthetic dataset they introduced. Su et al. \cite{su} proposed a multi-scale network working in combination with a discriminator module. The network has been trained combining three losses, one regarding the reconstruction performance, one enforcing a smoothness constraint, and an adversarial one. The architecture has then been tested on the synthetic dataset they introduced. Another dataset was introduced by Guo et al \cite{flat}, together with a deep learning model able to tackle both MPI and shot noise, and that is able to handle dynamic scenes too. Their model consists of an encoder-decoder architecture combined with a kernel prediction network used to tackle the shot noise.
Agresti et al. \cite{ag_2018} observed that the information regarding the structure of the scene is particularly important for MPI correction and, in order to have a simple network with about $150k$ parameters, they built it with two branches, one capturing the details and the other focusing on the high level geometry of the image. A similar idea was employed in \cite{dong} where a pyramid network observes the MPI structure at multiple resolutions, putting then the information together for the final prediction. In \cite{ag_unsup} the authors aimed at filling the gap between prediction on synthetic and real data, using an unsupervised domain adaptation approach. They took the model from \cite{ag_2018} and trained it as a GAN on unlabelled real data, clearly outperforming the original approach. The idea was later expanded by the same authors in \cite{ag_pami}, where they examined the possibility of performing domain adaptation also at input and feature level.
More recently, a few works tackled MPI correction making use of a more or less refined version of the light transport model. Barragan et al. \cite{itof2dtof} worked on the Fourier domain, using a U-Net architecture that takes a two-frequency input and predicts MPI corrupted data at several frequencies. They then compute the inverse Fourier transform on the output data, perform some filtering and get the depth prediction using a peak finding algorithm. The method shows good shot noise and MPI denoising capabilities but is quite heavy, with around $1.8M$ parameters, differently from the architecture proposed here, that works in the time domain and is much lighter.
In \cite{buratto} we  encoded the transient information with two peaks, the first one corresponding to the shortest light path, and the other to a weighted average of all other reflected light components. Even with such a rough approximation, we showed promising results on MPI correction on real data, all while using a network with a $3\times3$ receptive field. The method we propose now goes into the same direction as this previous work, since we focus more on the information on the transient dimension for the reconstruction rather than that on the spatial one. Apart from this high level similarity, the approach proposed here differs from the previous one in several aspects. First of all, we build a different learning architecture, which directly reflects the dual structure of transient information and at the same time employs a module that helps dealing with shot noise. We also introduce a more accurate encoding of the light transient information, and construct a specific network for its prediction.

In the literature, works directly targeted at transient recovery from iToF information are very few, all focusing on strong simplifying assumption for their solutions. Heide et al. \cite{heide} used an iToF camera to recover the depth information of a scene using the light reflected by a diffuse surface. They treated transient recovery as an optimization problem, constrained their solution both regarding spatial gradients and height field and introduced an algorithm in order to solve it. Lin et al. \cite{lin}, showed that the information recovered from a multi-frequency iToF camera corresponds to the Fourier transform of a transient image. They then proposed an algorithm for transient reconstruction from a high number of iToF modulation frequencies. On a different note, Liang et al. \cite{liang_compression} devised a deep learning model for the compression of rendered transient data, an important task due to the high volume of the data and the dangerous amount of rendering noise. 

%% file: sections/model.tex
\section{iToF Model}
\label{sec:model}

\begin{figure}[!h]
	\centering
	\includegraphics[width=0.8\columnwidth]{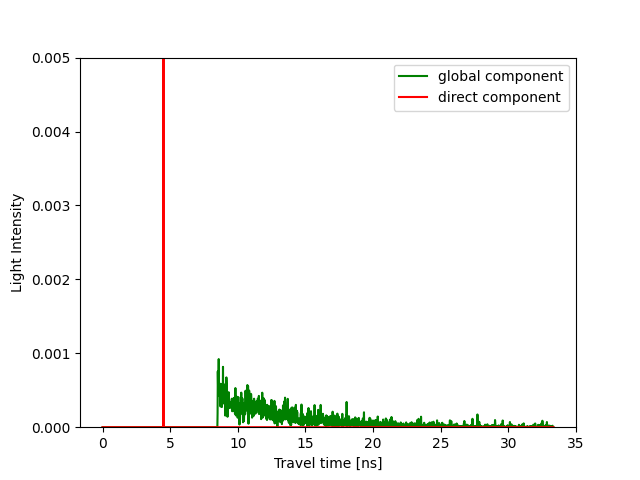}
	\caption{A sample transient vector from a scene in the \textit{walls} dataset. We highlighted the direct component in red and the global in green.}
	\label{transient}
\end{figure}

IToF cameras consist of an emitter and a sensor. The light sent by the emitter is a modulated signal $i(t)$, typically a sinusoidal wave with a frequency in the range of $10-100$ MHz, while the sensor function $s(t)$ is instead a periodic square wave with the same frequency. The iToF measurements are the result of the correlation between the reflected light signal $r(t)$ and $s(t)$,
\begin{equation}
	v_{\theta} = \int_0^{T_{int}} r(t)s\Big( t+\frac{\theta}{2\pi f_m}\Big) \text{d}t\mathrm{,}
	\label{eq:correlation}
\end{equation}
where $\theta$ is an internal phase shift, $v_{\theta}$ is the iToF measurement, $T_{int}$ is the integration time and $f_m$ the modulation frequency. In the ideal scenario where the reflected signal corresponds to a single light reflection, we have an analytical solution of the integral as $v_{\theta} = I + A \cdot \cos(\varphi+\theta)$, with $I$ the intensity of the signal, $A$ the amplitude of the sinusoid and $\varphi$ the phase delay due to the travel time. 
In a practical scenario 4 measurements are sufficient for recovering the $3$ unknowns, and from the phase delay $\varphi$ we are then able to reconstruct the depth information $d$ from the well-known relation:
\begin{equation}
	d = \frac{c \varphi}{4\pi f_m}\mathrm{,}
	\label{eq:depth}
\end{equation}
where $c$ is the speed of light.
If we take aside the intensity component, we can use the phasor notation as described in Gupta et al. \cite{gupta} to represent the iToF measurement with
\begin{equation}
	v = Ae^{i\varphi} = Ae^{i2\pi f_m \Delta t} \in \mathbb{C} \mathrm{,}
	\label{eq:phasor}
\end{equation} 
where $\Delta t$ is the round trip time of the light signal.
This notation can be used to mathematically describe the MPI phenomenon \cite{gupta} in a real case where the light bounces multiple times inside a scene, meaning that the sensor will receive and integrate not one but multiple signals covering different, normally longer, paths. This effect can now be written as follows, as phasors are closed under summation:
\begin{equation}
	v = \int_{t_{min}}^{t_{max}} x(t)e^{i2\pi f_m t} \text{d}t\mathrm{,}
	\label{eq:MPI_int}
\end{equation}
with $t_{min}$ and $t_{max}$ respectively the minimum and maximum time of flights considered, and $x(t)$ the time dependent scene impulse response, also known as transient. A discretized transient can be seen in Figure \ref{transient}.
We can now discretize the integral in Equation (\ref{eq:MPI_int}), 
\begin{equation}
	v = \sum_{t=t_{min}}^{t_{max}} x(t)e^{i2\pi f_m t} \mathrm{,}
	\label{eq:MPI_sum}
\end{equation}
consider multiple acquisition frequencies and rewrite it as 
\begin{equation}
	\boldsymbol{v}  = \boldsymbol{\Phi} \boldsymbol{x}\mathrm{,}
	\label{eq:MPI}
\end{equation}
\\where $\boldsymbol{v}$ are the iToF measurements at multiple frequencies, $\boldsymbol{x}$ is the scene impulse response and $\boldsymbol{\Phi}$ the measurement model linked to the iToF camera.

%% file: sections/method.tex
\section{Method}
\label{sec:method}

In this section, we provide an exhaustive description of a novel method for MPI correction and transient image reconstruction. We start by describing the idea behind it and then go in detail through each of the three components of our modular architecture: the \textit{Spatial Feature Extractor}, the \textit{Direct Phasor Estimator} and the \textit{Transient Reconstruction Module}. In the end of the section, we introduce the losses employed for training.

\subsection{Direct-Global Subdivision}

Let's begin by considering the structure of a common transient vector (see the example in Figure \ref{transient}); it is quite clear that it is composed of two quite distinct parts: one corresponding to the first peak, the other instead incorporating all the other incoming light rays.  
From now on, we will call the vector composed by the first peak alone \textit{direct component} and will denote it with $\boldsymbol{x_d}$, while the \textit{global component} will be composed of all the other reflections and will be represented as $\boldsymbol{x_g}$.
We can now consider Equation (\ref{eq:MPI}) and write
\begin{equation}
\label{eq:directglobal}
\boldsymbol{v} = \boldsymbol{\Phi}\boldsymbol{x} = \boldsymbol{\Phi}(\boldsymbol{x_d}+\boldsymbol{x_g}) = \boldsymbol{v_d}+\boldsymbol{v_g} \mathrm{,}
\end{equation}
where we exploited the linearity of the model to extract the $\boldsymbol{v_d}$ and $\boldsymbol{v_g}$ vectors.
What follows from this derivation is that the subdivision of the transient vector into direct and global components can be translated also onto the iToF domain.
In practice, we now have a vector $\boldsymbol{v_d}$ which corresponds to ideal iToF measurements, the ones that would be produced by the direct peak alone, while $\boldsymbol{v_g}$ are the measurements corresponding to all reflections but the first.
\\

\begin{figure}[!h]
	\centering
	\includegraphics[width=0.99\columnwidth]{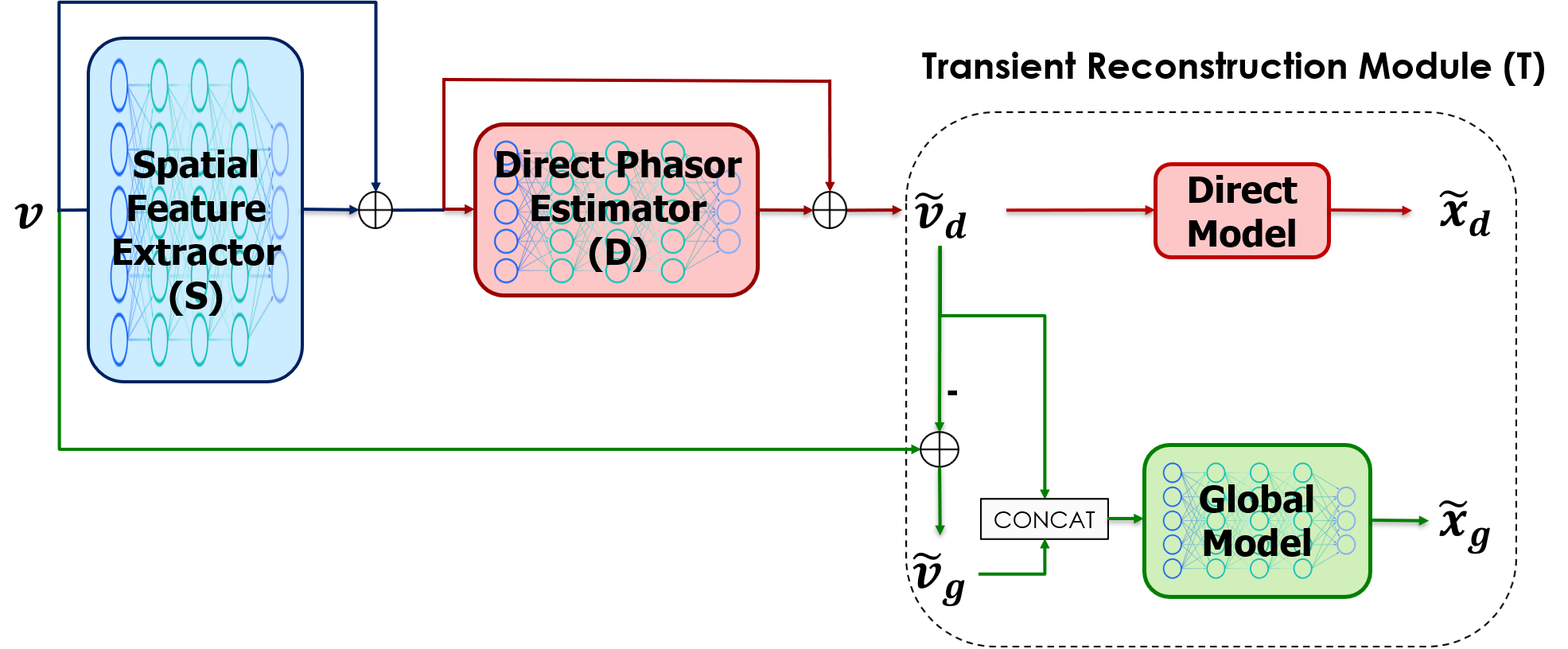}
	\caption{High level structure of our training architecture}
	\label{model}
\end{figure}
\begin{figure*}[t]
	\centering
	\includegraphics[width=1.6\columnwidth]{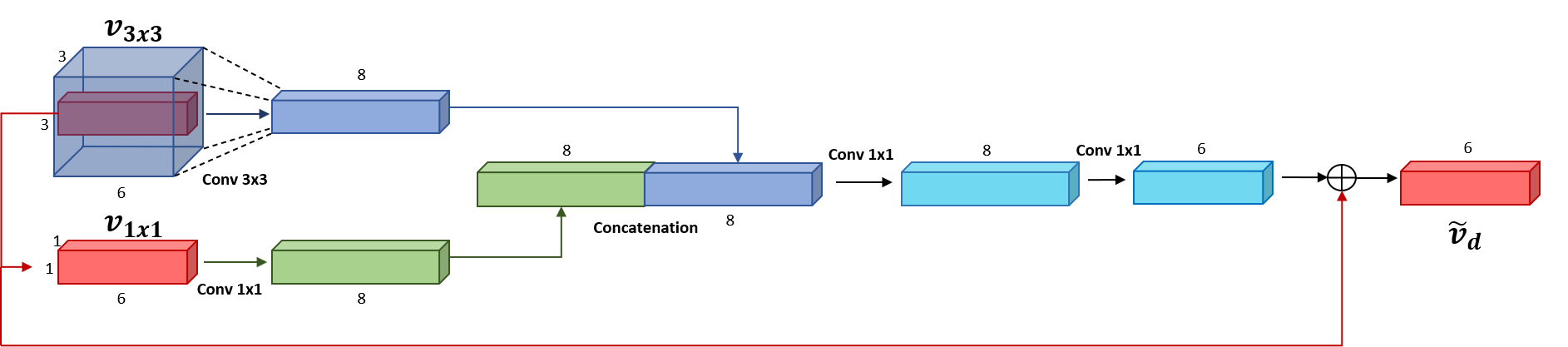}
	\caption{Representation of the Direct Phasor Estimator module for $3$ input frequencies and $8$ feature maps}
	\label{fig:directcnn}
\end{figure*}

\subsection{Deep Learning Architecture}
The different modules of our network are shown in Figure \ref{model}. As input the model takes in the real and imaginary components of the raw iToF measurements $\boldsymbol{v}$ at different modulation frequencies. First, they go through the \textit{Spatial Feature Extractor}, which exploits the spatial information to produce an intermediate representation of the data (it proved to be very useful for handling zero-mean noise). Its output is then processed by the \textit{Direct Phasor Estimator}, that predicts the iToF measurements corresponding to the direct component, which, subtracted from the original input, gives us also the iToF measurements corresponding to the global component. The two predictions are in the end fed to the \textit{Transient Reconstruction Module} that has the task of reconstructing the whole transient vector. As we will see, this module is further split into the \textit{Direct Model} which is a deterministic function computing the direct component, and the \textit{Global Model} that instead consists of a deep learning architecture predicting the global component. 
For the construction of the learning model we used as a starting point the network introduced in \cite{buratto}, where the raw iToF input is directly mapped into an oversimplified encoded version of a transient vector, consisting of two peaks. We kept a narrow receptive field claiming that the information in the transient dimension is enough for MPI correction. Differently from \cite{buratto}, we introduce a middle step between the input $\mathbf{v}$ and the transient prediction $\mathbf{\tilde{x}}$ (i.e. the subdivision between direct and global components), and provided a more complex and realistic model for the backscattering vector itself.

\subsubsection{Spatial Feature Extractor (S)} The main task of this module is providing an encoded version of spatial information to the following stages. As we will see, the \textit{Direct Phasor Estimator} has a very narrow receptive field (i.e. $3\times 3$), which limits its capability of managing noise sources such as shot noise. The \textit{Spatial Feature Extractor} is a fully convolutional architecture with a $9\times 9$ receptive field. It consists of $4$ layers, each with $32$ feature maps and a residual connection links the central $3\times3$ part of the input to the output.
\subsubsection{Direct Phasor Estimator (D)} This module estimates $\boldsymbol{v_d}$, the direct component of the raw phasor. The raw measurements coming from the \textit{Spatial Feature Extractor} are fed to two branches with receptive field $3\times3$ and $1\times1$ respectively, whose outputs are then concatenated and used for the prediction of $\boldsymbol{\tilde{v}_d}$. 
More in detail, as depicted in Figure \ref{fig:directcnn}, it takes in input both a $3\times3$ patch and its central pixel; they go through a convolutional layer with an output of size $1\times1$ and are then concatenated. The information is then processed by two other convolutional layers before producing the $\boldsymbol{\tilde{v}_d}$ prediction. Each convolutional layer has a total of $8$ feature maps and there is a residual connection between input and output.
From the prediction of $\boldsymbol{\tilde{v}_d}$ we then compute the corresponding depth for each of the modulation frequencies, using the smallest frequency for solving ambiguity range uncertainty on the higher ones. The output depth maps are then passed through a bilateral filter and the final depth prediction will be the pixel-wise minimum of the output depths. The reason for this is that the MPI, that is the major cause of error, leads  to an overestimation of the distance.
Considering Equation (\ref{eq:directglobal}) we can then retrieve also the iToF measurements corresponding to the global component by simply subtracting the direct component, i.e., $\boldsymbol{\tilde{v}_g}=\boldsymbol{v}-\boldsymbol{\tilde{v}_d}$. 
Notice that this module is tackling the MPI removal task, as if the direct-global subdivision is successful, we are able to recover an MPI-free estimate of our input from the $\boldsymbol{\tilde{v}_d}$ component. A visual representation of this module can be found in Figure \ref{fig:directcnn}.

\subsubsection{Transient Reconstruction Module (T)}
Retrieving the transient information from iToF leads to some serious challenges, not only linked to the difficulty of the task itself, but also to the dimensionality of our output. As remarked in Section \ref{sec:introduction}, we want to map the raw iToF measurements, that correspond to a handful of values, into a vector with thousands of entries. The complexity of the matter makes an encoding of the ground truth a necessity. Since the one proposed in \cite{buratto} is way too simplistic, and the one by Liang et al. \cite{transient_compression} computationally heavy, we propose a novel approximation of the transient vector $\boldsymbol{x_g}$ with just 6 parameters, 2 needed for the direct component, and the other 4 for the global.
Therefore, the \textit{Transient Reconstruction Module} is further split into two components, the \textit{Direct Model} that takes care of the reconstruction of the direct component and the \textit{Global Model}, which instead predicts the global component.
\paragraph{Direct Model}
Similarly to \cite{buratto}, each direct component $\boldsymbol{x_d}$ gets encoded by its magnitude $E_d$ and time position $t_d$. 
As a matter of fact, no learnable parameters are needed for the prediction of the direct component $\boldsymbol{x_d}$, since the time position $t_d$ is directly proportional to the phase $\varphi_d$ through Equation (\ref{eq:phasor}), and can be retrieved directly from $\boldsymbol{\tilde{v}_d}$. At the same time, the magnitude of the first peak $E_d$ is strictly related to the amplitude of the raw iToF measurements of the direct component; this is true since in the case of a single peak, the magnitude is the value of the peak itself, while the amplitude $A_d$ can be written as follows,
\begin{align}
	A_d &= \frac{1}{2}\sqrt{\boldsymbol{v_{d,\Re}}^2+\boldsymbol{v_{d,\Im}}^2} \nonumber
	\\ &= \frac{1}{2}\sqrt{\left(\sum_{t=0}^T \Phi_{\Re,t}\boldsymbol{x_t}\right)^2+\left(\sum_{t=0}^T \Phi_{\Im,t}\boldsymbol{x_t}\right)^2} =
	\\ &= \frac{1}{2} \sqrt{\left(\Phi_{\Re,t_d} x_{t_d}\right)^2  + \left(\Phi_{\Im,t_d} x_{t_d}\right)^2} =\nonumber 
	\\ &= \frac{1}{2}x_{t_d} =\frac{1}{2} E_d\nonumber  \mathrm{,}
	\label{eq:first_peak_magnitude}
\end{align}
where we used the Pythagorean identity together with the fact that only one element of the sum is non-zero (the one at time index $t_d$). $\boldsymbol{v_{d,\Re}}$ and $\boldsymbol{v_{d,\Im}}$ are the real and imaginary components following the phasor notation in Equation (\ref{eq:phasor}). Given this derivation, we need no additional learning parameters for the \textit{Direct Model} . 

\paragraph{Global Model}
For the encoding of $\boldsymbol{x_g}$ we chose instead the following parametric function $\boldsymbol{\tilde{x}_g}(t)$ inspired by the Weibull distribution \cite{weibull}
\begin{equation}
\label{eq:weibull}
\boldsymbol{\tilde{x}_g}(t) = \mathbf{a}(t-\mathbf{b})^{\mathbf{k}-1}\exp{\left( -\frac{t-\mathbf{b}}{\boldsymbol{\lambda}} \right) ^\mathbf{k}} \mathrm{,}
\end{equation}  
where $t$ ranges from $0$ to $T$ (the maximum acceptable travel time), $\mathbf{a}$ takes care of the scale, $\mathbf{b}$ of the shift, and $\mathbf{k}$ and $\boldsymbol{\lambda}$ of the shape. For the choice of this function we took inspiration from the topic of multipath interference related to radio signals, where distributions such as the Rayleigh or the Weibull are usually employed \cite{weibull}. In the end we decided to employ the Weibull distribution since it is a generalization of the Rayleigh and shows a good resemblance with common shapes of transient vectors.
Predicting the parameters of the global component $\boldsymbol{\tilde{x}_g}$ expressed in Equation (\ref{eq:weibull}) from $\boldsymbol{\tilde{v}_g}$ and  $\boldsymbol{\tilde{v}_d}$ is a quite complex task, which is handled by an additional deep learning architecture.  The \textit{Global Model} is composed of $4$ parallel branches with a $1\times1$ receptive field, each predicting one of the 4 parameters of the parametric function. Each branch is composed of a stack of $2$ convolutional layers with a total of $32$ feature maps each. It takes $\boldsymbol{\tilde{v}_g}$ and  $\boldsymbol{\tilde{v}_d}$ in input, estimates from it the 4 parameters of the function described in Equation (\ref{eq:weibull}), and finally compares it to the ground truth $\boldsymbol{x_g}$. 
\\Combining together the outputs of the \textit{Direct Model} and of the \textit{Global Model}, we obtain an estimate of the transient vector.
 \subsection{Training Targets}
 The losses used for training our architecture are the Mean Absolute Error (MAE), and the Earth Mover's Distance (EMD) \cite{emd}. The \textit{Direct Phasor Estimator} uses as guidance a simple MAE on the target values, while the EMD guides the training of the \textit{Global Model}.
 The ground truth $\boldsymbol{v_d}$ can only be retrieved from the transient information and for this reason it is not available when using common iToF datasets. For this reason, we employed two different training methodologies according to the input data:
 \begin{enumerate}
 	\item When the transient data is available we can directly compute the loss between the ground truth $\boldsymbol{v_d}$ and our prediction $\boldsymbol{\tilde{v}_d}$ as 
 	\begin{equation}
 		\mathcal{L}_{MAE_{v_d}} = \mathop{\mathbb{E}}\left[ \norm{\boldsymbol{v_{d}}-\boldsymbol{\tilde{v}_{d}}}_1\right] \mathrm{.} 		
 		\label{eq:maevd}
 	\end{equation}
    \item When instead the dataset only offers the depth ground truth, we are unable to recover $\boldsymbol{v_d}$, but we are able to compute the ground truth phase delay $\boldsymbol{\varphi_d}$ following Equation (\ref{eq:depth}). From the network prediction $\boldsymbol{\tilde{v}_d}$ we can thus compute the predicted $\boldsymbol{\tilde{\varphi}_d}$ through an arctangent operation, and finally compute the loss as: 
    \begin{equation}
    	\mathcal{L}_{MAE_{\varphi_d}} = \mathop{\mathbb{E}}\left[ \norm{ \boldsymbol{\varphi_{d}}-\boldsymbol{\tilde{\varphi}_{d}}}_1\right] \mathrm{.}
    	\label{eq:maephid}
    \end{equation}
 \end{enumerate} 
 Furthermore, if the training dataset instead contains not only the depth ground truth, but also images both with and without zero-mean noise, it is possible to extend the interpretability of the architecture, by pinning the output of the \textit{Spatial Feature Extractor} with an additional loss. The \textit{Spatial Feature Extractor} would therefore be dealing only with zero-mean noise, while the \textit{Direct Phasor Estimator} would take care exclusively of MPI. 
 The output of the \textit{Global model} is guided instead by the EMD, which we had already employed in \cite{buratto}, and is defined as
\begin{equation}
\label{eq:emd}
\mathcal{L}_{EMD} = \mathop{\mathbb{E}} \left[\norm{\boldsymbol{X_{g}} - \boldsymbol{\tilde{X}_{g}}}_1\right] \mathrm{,}
\end{equation}
where $\boldsymbol{X_g}$ and $\boldsymbol{\tilde{X}_g}$ are the cumulative sums of $\boldsymbol{x_g}$ and $\boldsymbol{\tilde{x}_g}$ respectively. What this distance measure captures is the dissimilarity between the two distributions, i.e., the minimum amount of work needed to convert one into the other \cite{emd}.
\\The performance of the different losses and the modularity of the approach will be thoroughly investigated in Section \ref{sec:results}.

%% file: sections/datasets.tex
\section{Datasets}
\label{sec:datasets}
In this section we will introduce the main datasets employed for the training and evaluation of our model. We will employ both iToF and transient datasets, due to the close relation between the two topics and the lack of datasets of the second kind. In particular, we will introduce the \textit{Walls} dataset: a novel synthetic transient dataset based on simple structures which will be used for training and for evaluating the $\textit{Transient Reconstruction Module}$.

\subsection{iToF Datasets}
Regarding the iToF data, we will mainly focus on the synthetic and real datasets introduced in \cite{ag_2018,ag_unsup}. These datasets come with amplitude and phase information at three different modulation frequencies: $20,50$ and $60$ MHz; the scenes depicted are simple indoor scenes, with a maximum distance smaller than $7.5$ \textit{m} (the ambiguity range of the $20$ MHz component) and high amounts of MPI. In particular, the synthetic dataset $S_1$ is composed of $54$ scenes ($40$ for training and $14$ for testing), has a high degree of shot noise and a spatial resolution of $240\times320$, while $S_3$, $S_4$ and $S_5$ are all real datasets with $8$ images each, a limited amount of shot noise and a spatial resolution of $239\times320$. Dataset $S_1$ will be employed for training, $S_3$ for validation and $S_4$ and $S_5$ will be the main test sets for benchmarking the MPI correction capabilities of our network.
The iToF2dToF dataset \cite{itof2dtof} will instead be used for some additional studies regarding the resilience to shot noise and MPI correction. The dataset is composed of a total of $5000$ images with a spatial resolution of $120\times160$ and with all iToF measurements ranging from $20$ to $600$ MHz with a step of $20$. The dataset presents an extremely high amount of shot noise (around $80\%$ of the total noise) and will be used as a stress test for our training architecture.

\subsection{The \textit{Walls} Transient Dataset}
\begin{figure}[!h]
\centering
\subfloat[Two walls]{\includegraphics[width=0.5\columnwidth,trim=6ex 6ex 1.5ex 6ex,clip]{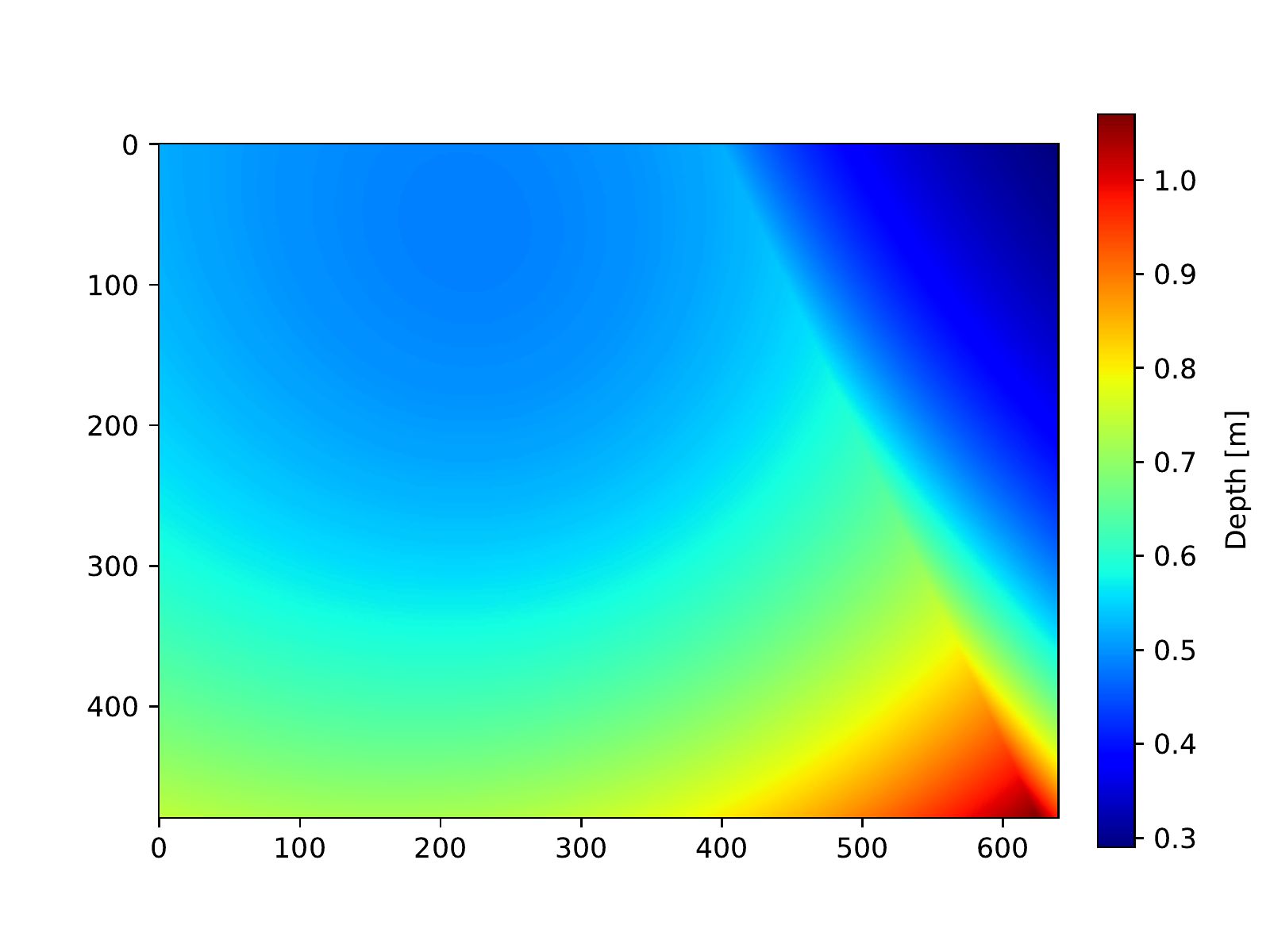}
\label{twalls}}
\subfloat[Three walls]{\includegraphics[width=0.5\columnwidth,trim=6ex 6ex 1.5ex 6ex,clip]{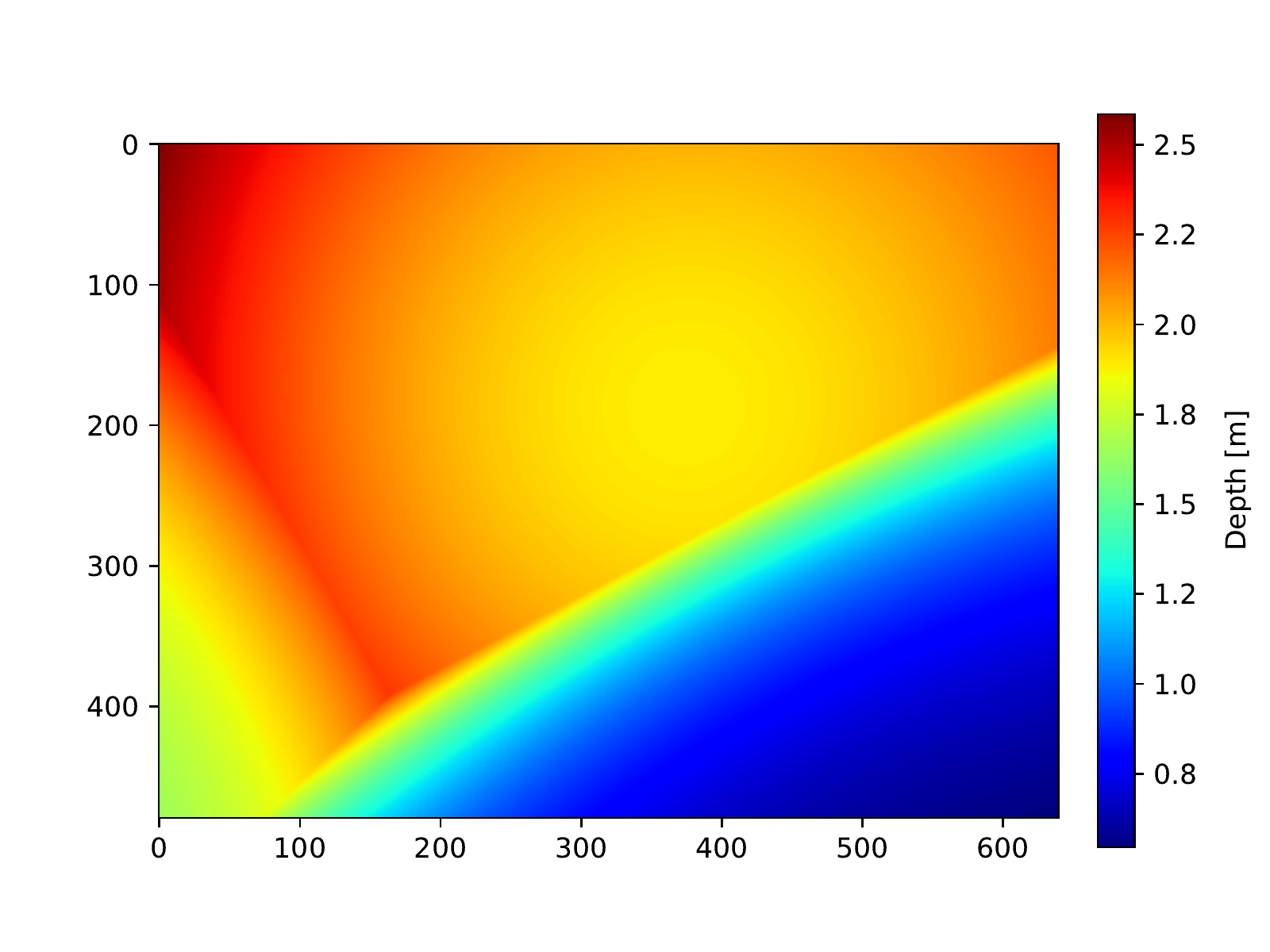}
\label{3walls}}
\caption{Depth images from our transient dataset}
\label{walls}
\end{figure}

The task of transient reconstruction is quite new in the literature and this is also due to the difficulties in acquiring a reliable transient dataset for the task. No real transient datasets are available and only few synthetic ones are freely accessible such as the FLAT dataset \cite{flat} and the Zaragoza \cite{transient_dataset} one. 
\\In this paper we introduce the \textit{Walls} dataset: a novel synthetic transient dataset based on simple geometries. The dataset has been simulated using the Microsoft ToF Tracer \cite{toftracer} with a maximum depth set to $5$ \textit{m}. The simulated scenes consist of one to three walls with varying angles between them. The point is that the dataset has been built as a template case for MPI. The scenes, while very simple, still capture some of the most common MPI scenarios where the overestimation is due to at maximum a couple of reflecting surfaces. This assumption may not be true in general, but it is a good approximation for most practical cases, as the light intensity is inversely proportional to the square of the travelled distance, making the contributions of longer paths mostly negligible. In total, the dataset is composed of $222$ images, $53$ with a single wall, $95$ with two, and $74$ with three. A couple of samples can be seen in Figure \ref{walls}.
The spatial resolution of our images has been set to of $480\times640$, to match that of some of the most recent ToF cameras, while the temporal dimension has been divided into $2000$ bins; keeping into account that the maximum depth is $5$ \textit{m}, this means that the depth quantization step consists of $2.5$ \textit{mm}, a desirable property for indoor settings.
The dataset has no noise sources other than MPI and rendering noise. 

As the maximum depth of the \textit{Walls} dataset amounts to $5$ meters, while the one of $S_1$ gets to $7.5$ meters, we decided to perform some data augmentation on the \textit{Walls} dataset in order to cover the wider range. In practice, we added a shift to each of the transient vectors, randomly picking it from a uniform distribution in the $\left[0 , 5\right] m$ range and from these we then recomputed the iToF measurements. The dataset can be found at the following \href{https://lttm.dei.unipd.it/paper_data/transientMPI/}{link}.

In the following Section we will benchmark our model for the task of MPI correction and provide qualitative results both for this task and for the one of transient reconstruction.

%% file: sections/results.tex
\section{Results}
\label{sec:results}
This section is devoted to the experimental evaluation of our method and to the comparison with the existing state-of-the-art. The quantitative evaluation and comparison will be carried out on the MPI correction task, while the transient reconstruction part will be evaluated only qualitatively.

\subsection{Training Details}
The proposed neural networks have been trained using the $S_1$ and \textit{Walls} datasets. From the first one we took the original training set of $40$ images, while from the \textit{Walls} dataset we randomly picked $134$ images. The validation set consists instead of the 8 images from the real dataset $S_3$. In particular, the training data has been cut in patches of size $11\times11$, randomly chosen inside the images, while the validation set has been kept at full resolution. 
The models have been developed in Tensorflow $2.1$, the trainings of our architecture have been performed on an NVIDIA 2080 Ti GPU, with ADAM as optimizer with a learning rate of $10^{-4}$ and a batch size of 2048. 
We will focus our evaluation for MPI correction on two models: the first one comprised of the first two modules introduced in Section \ref{sec:method} which we will abbreviate \textit{SD}, and the second a lighter architecture without the \textit{Spatial Feature Extractor}. In this case the number of feature maps of the \textit{Direct Phasor Estimator} was changed from $8$ to $32$ to provide the network with additional learning parameters. We will abbreviate this second model \textit{D}.  
In all cases, each input patch has been normalized by the mean amplitude of its $20$ MHz component to help generalizing on real data.

\subsection{Results on MPI correction}
We will now compare our approach with some of the best performing MPI correction methods. The comparison will be made with SRA \cite{sra}, an algorithmic approach, with DeepToF \cite{deeptof}, one of the first deep learning approaches for MPI correction, with Buratto et al. \cite{buratto}, which was the starting point for our current architecture and with the approach from Agresti et al. \cite{ag_2018}, together with the subsequent domain adaptation approaches \textit{in}-DA, \textit{feat}-DA and \textit{out}-DA proposed in \cite{ag_pami}. 

In Table \ref{tab:results} we show the overall comparison between the cited approaches and the two proposed architectures. The first two columns show the MAE on the two real datasets $S_4$ and $S_5$, while the last one shows the network complexity of each approach; SRA has no entry as it isn't deep learning based. Regarding our approaches, the parameters of the \textit{Transient Reconstruction Network} were not included in the total amount as it is not needed for MPI correction. Moreover, note that while the \textit{SD} model has been trained on both the $S_1$ and \textit{Walls} datasets, \textit{D} has been trained on \textit{Walls} alone, since as we will see it is not able to deal with shot noise due to its very narrow receptive field.
The real datasets present a clear challenge due to the domain shift between synthetic and real data. In practice, the resemblance between training and test data is strictly limited by the accuracy of the simulation, which can mimic a real scenario only up to a certain extent. As we can see from the table, our approach not only clearly outperforms both the architectures proposed in \cite{buratto} and \cite{ag_2018}, but also beats the results from \cite{ag_pami} on $S_4$, while falling shortly behind on $S_5$, all by using just $\frac{1}{7}$ of its parameters. This is particularly striking as differently from \cite{ag_pami} we only rely on synthetic data for our prediction. Given this, we have clear reasons to expect an even better performance by using some unsupervised domain adaptation techniques as was done in \cite{ag_pami}.   Another interesting outcome is the fact that the \textit{D} model shows quite competitive results w.r.t. \textit{SD} and \cite{ag_pami} and at the same time outperforms other architectures such as \cite{ag_2018} and \cite{buratto}. The \textit{D} model is extremely light, with just around $3$k learnable parameters, but still gets close to state-of-the-art results. 

\begin{figure*}
	\centering
	\begin{tabular}{ccccc}
		 1 frequency (60 MHz) & DeepToF \cite{deeptof} & Buratto et al. \cite{buratto} & \textit{out}-DA \cite{ag_pami} & Ours: SDT\\
		 \includegraphics[width=2.5cm,trim=21ex 8ex 18ex 11.2ex,clip]{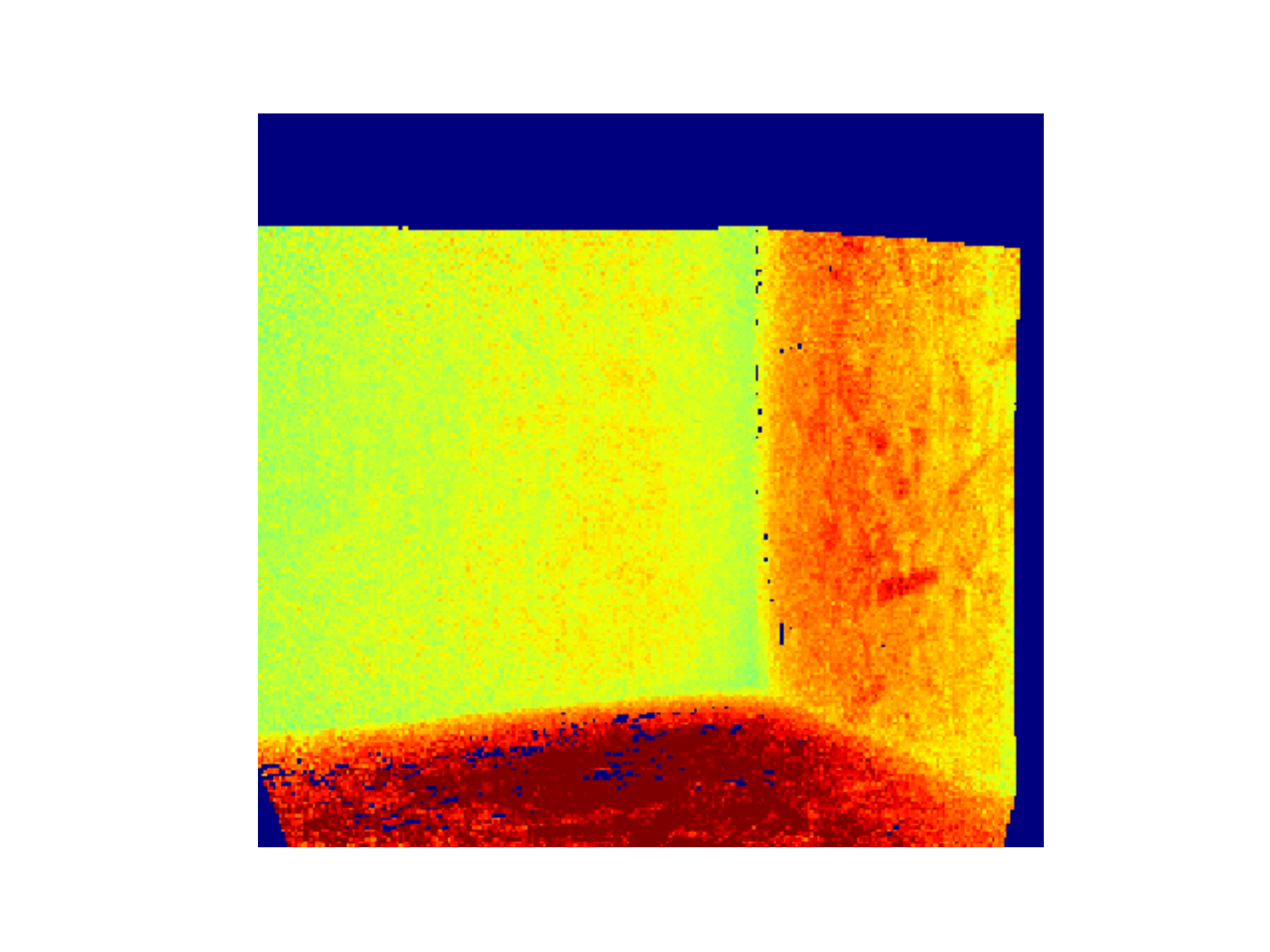}&		 \includegraphics[width=2.5cm,trim=0ex 2ex 15.5ex 2ex,clip]{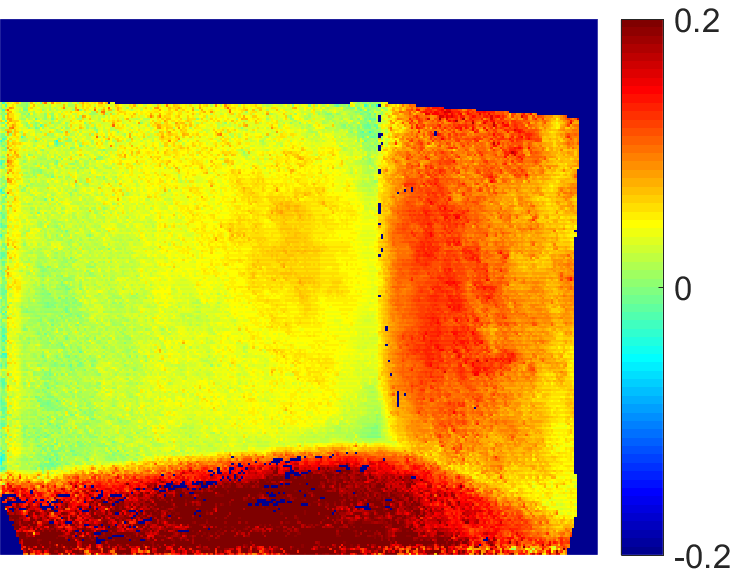}&		 \includegraphics[width=2.5cm,trim=21.7ex 8ex 21ex 7ex,clip]{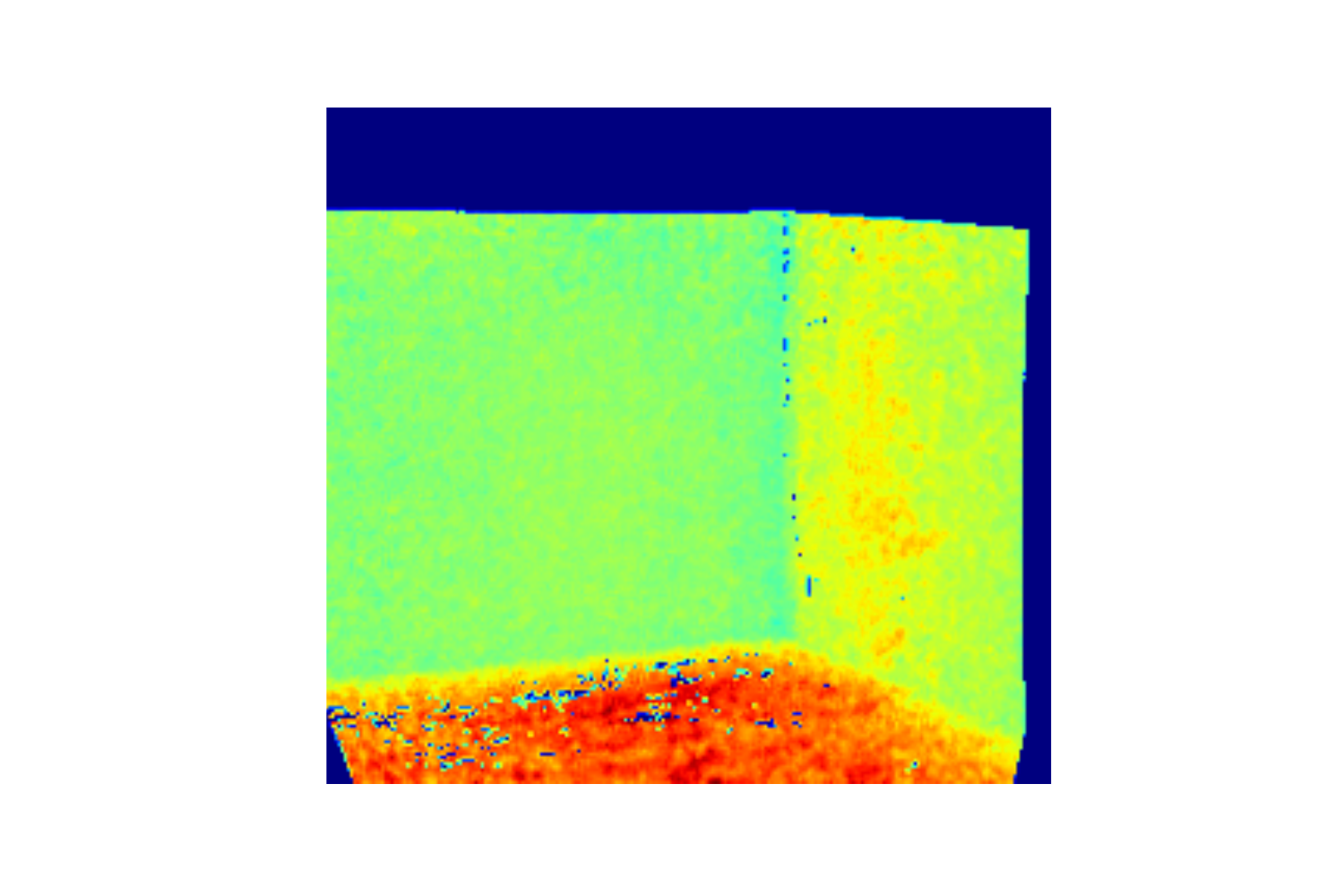}&		 \includegraphics[width=2.5cm,trim=0ex 2ex 15.5ex 2ex,clip]{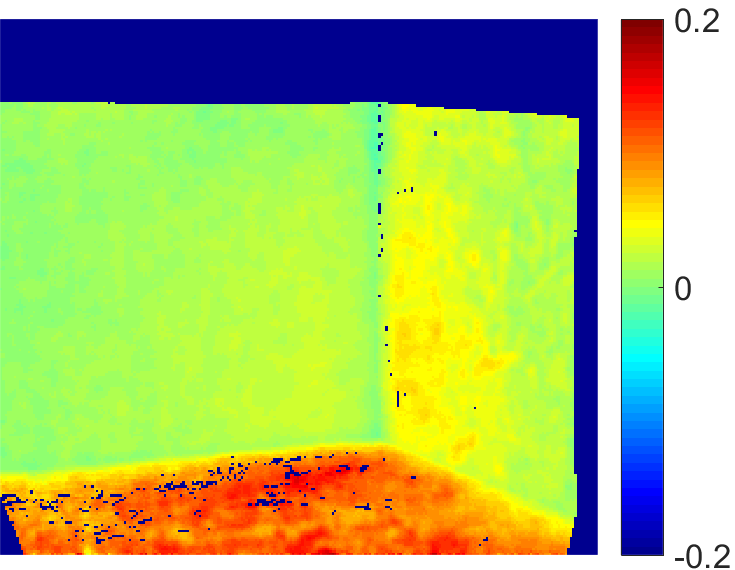}&		 \includegraphics[width=3cm,trim=22ex 8ex 1ex 4ex,clip]{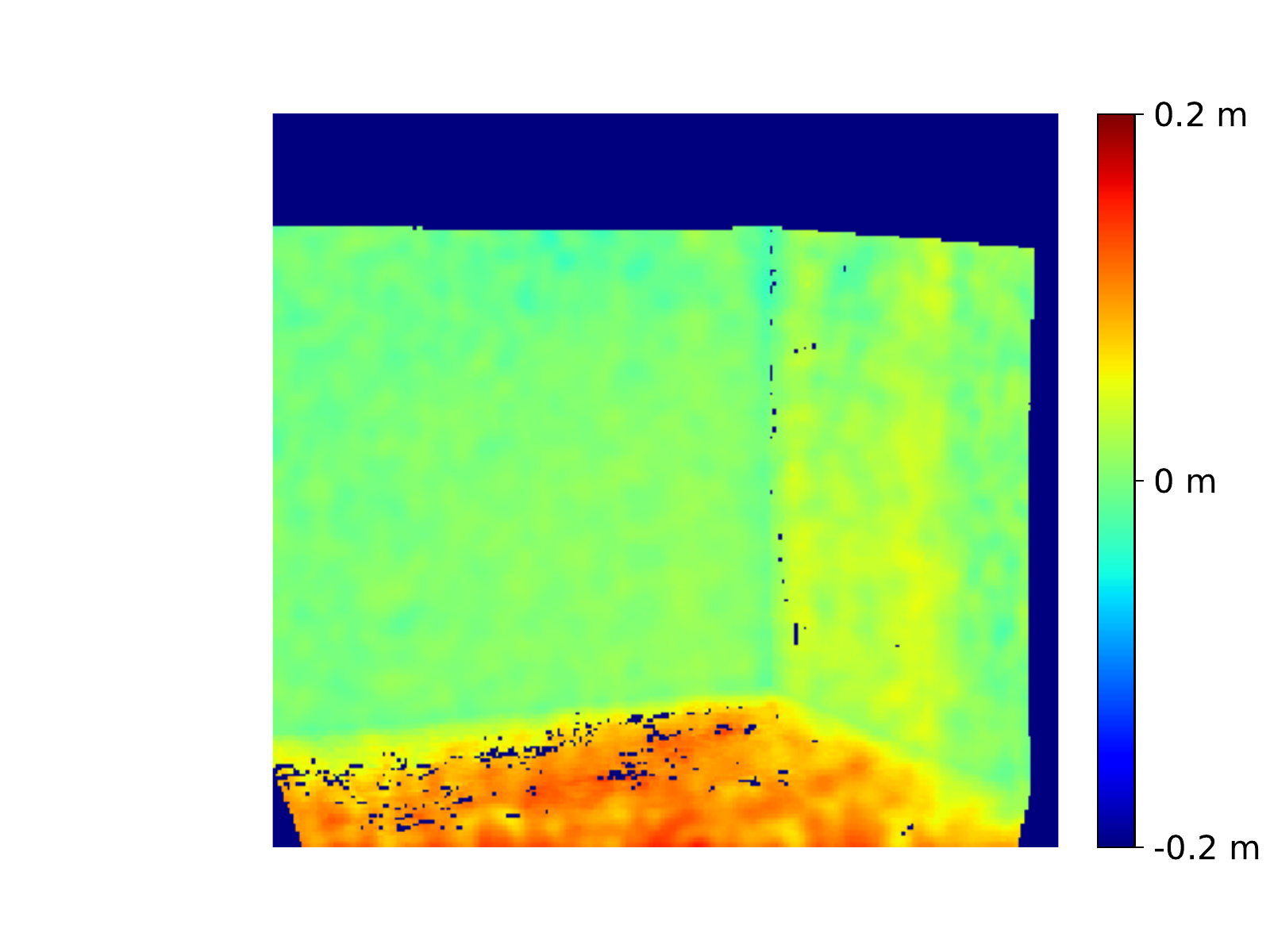}\\
		\includegraphics[width=2.5cm,trim=21ex 8ex 18ex 11.2ex,clip]{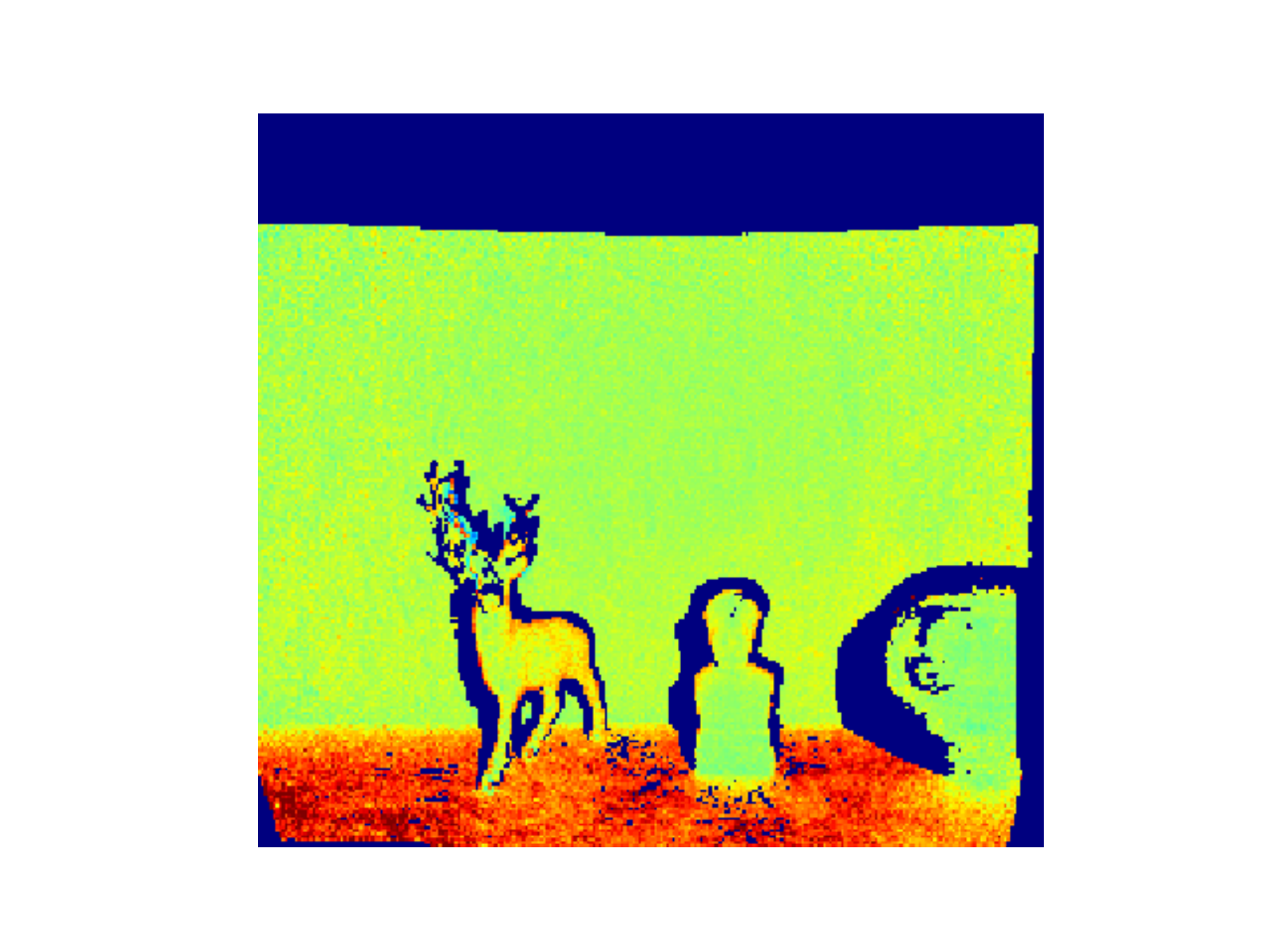}&		 \includegraphics[width=2.5cm,trim=0ex 2ex 15.5ex 2ex,clip]{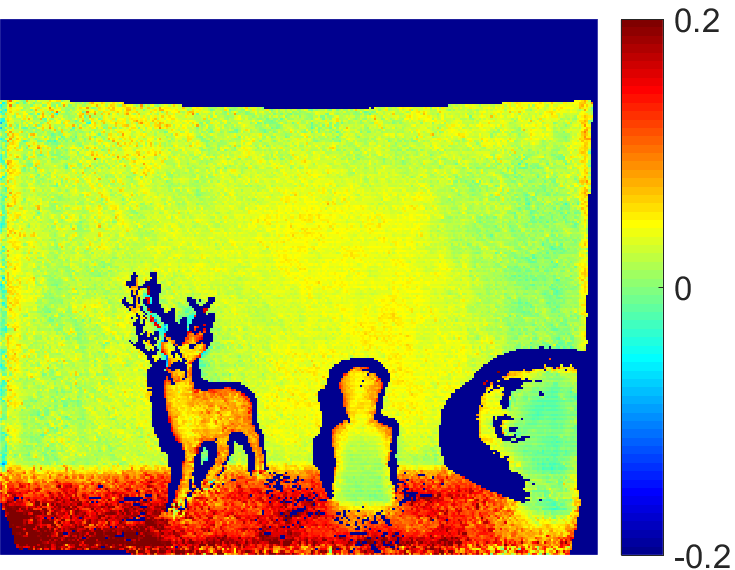}&		 \includegraphics[width=2.5cm,trim=21.7ex 8ex 21ex 7ex,clip]{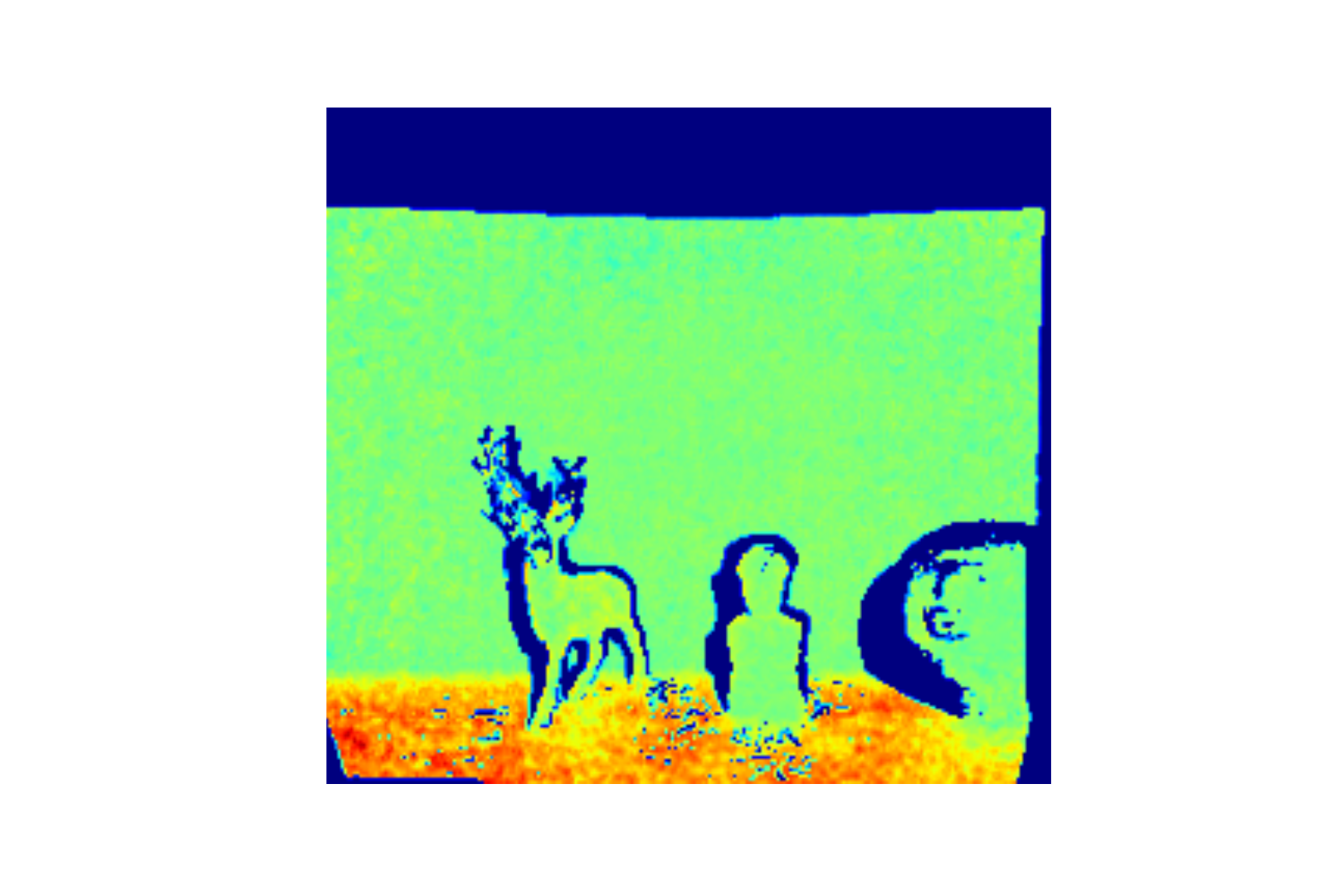}&		 \includegraphics[width=2.5cm,trim=0ex 2ex 15.5ex 2ex,clip]{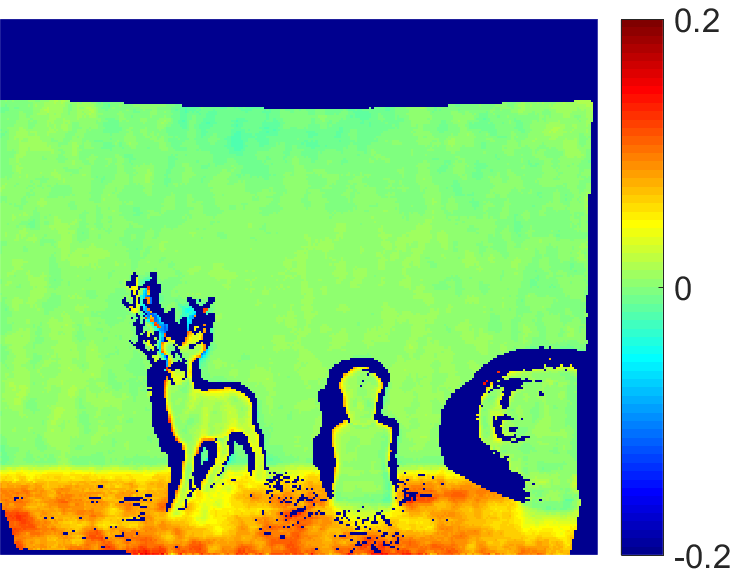}&		 \includegraphics[width=3cm,trim=22ex 8ex 1ex 4ex,clip]{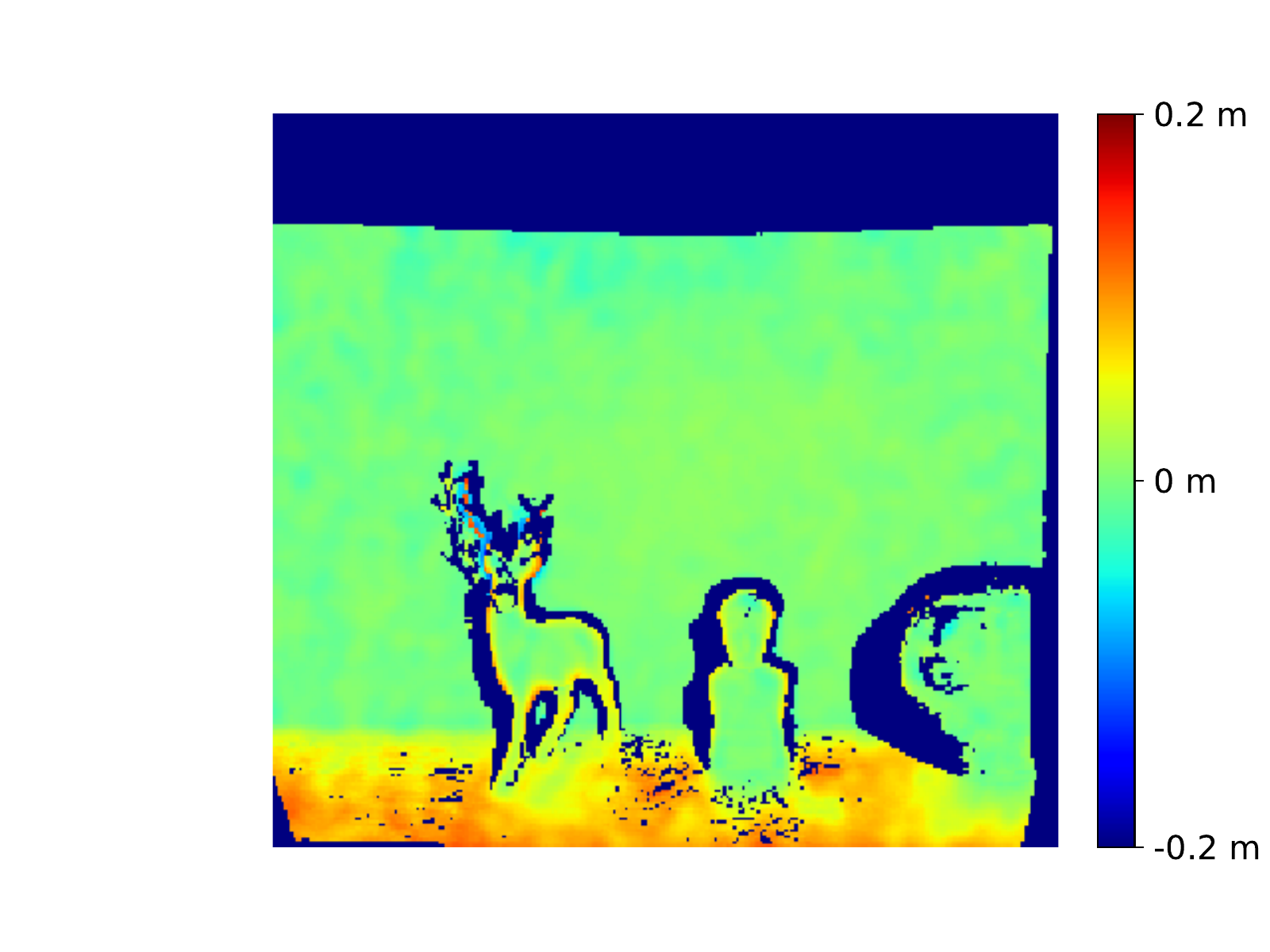}\\	
		\includegraphics[width=2.5cm,trim=21ex 8ex 18ex 11.2ex,clip]{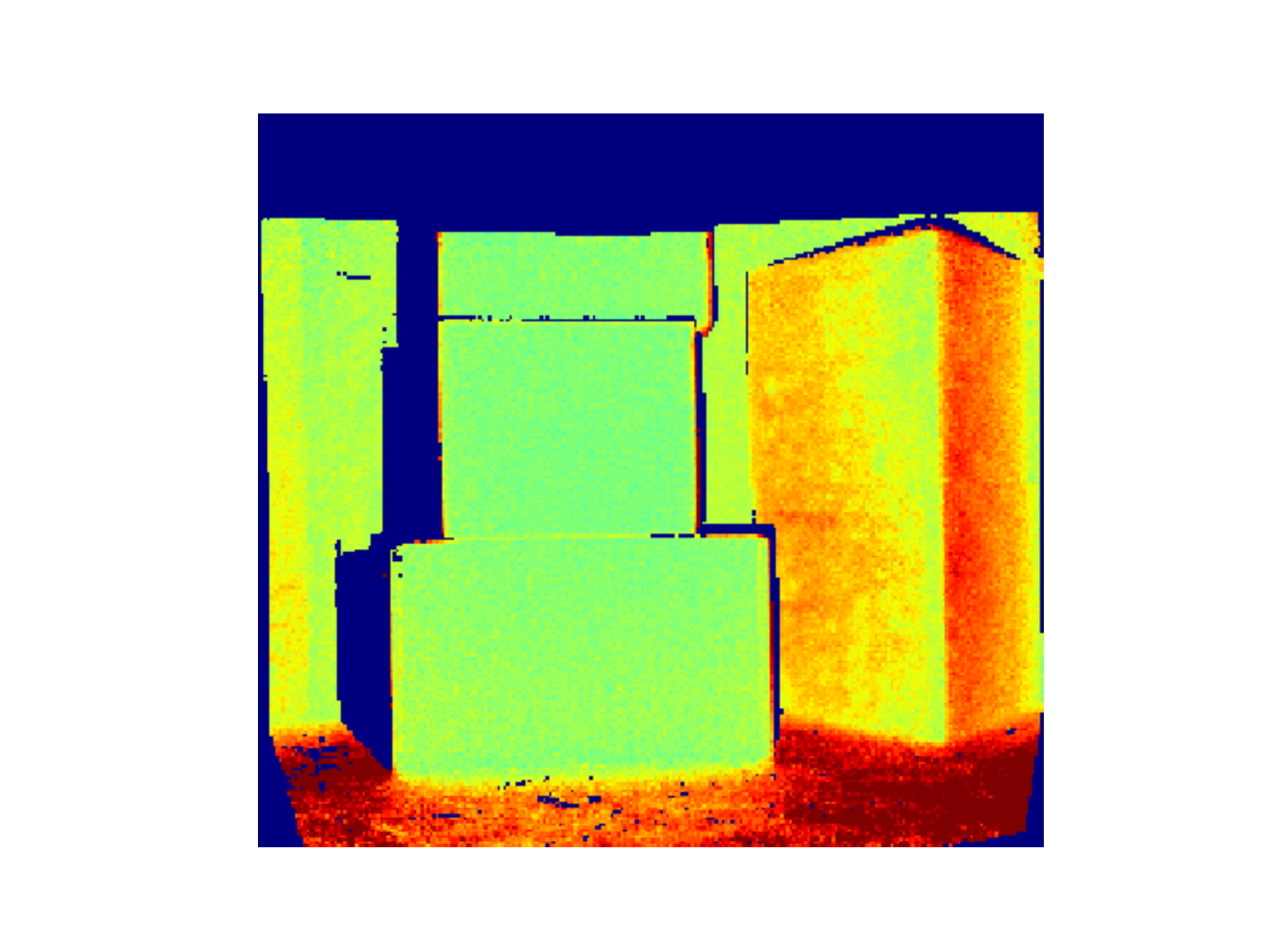}&		 \includegraphics[width=2.5cm,trim=0ex 2ex 15.5ex 2ex,clip]{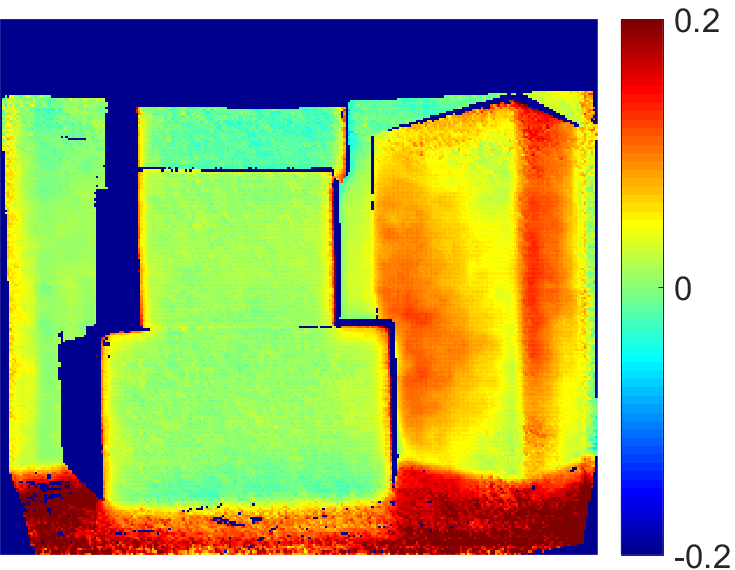}&		 \includegraphics[width=2.5cm,trim=21.7ex 8ex 21ex 7ex,clip]{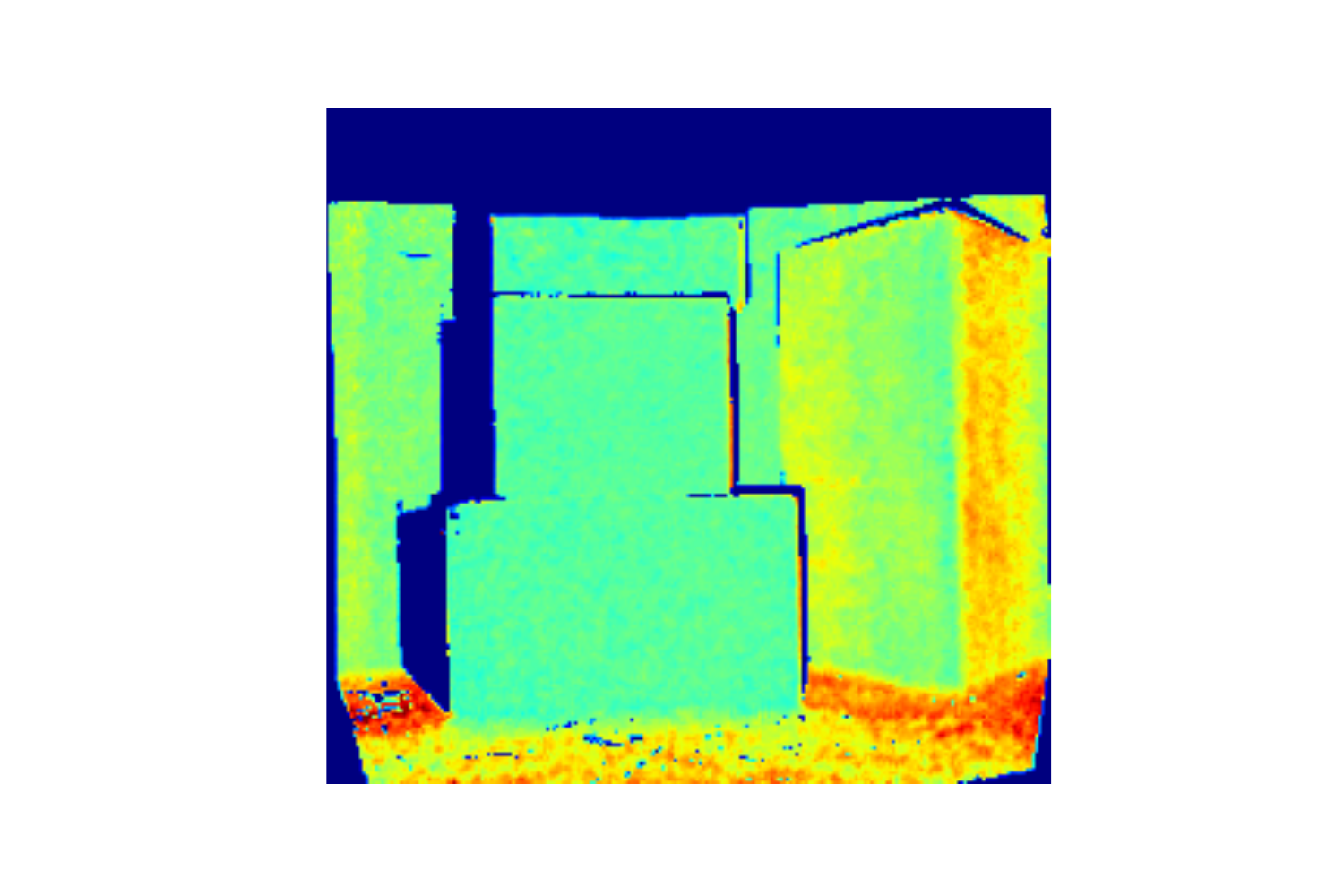}&		 \includegraphics[width=2.5cm,trim=0ex 2ex 15.5ex 2ex,clip]{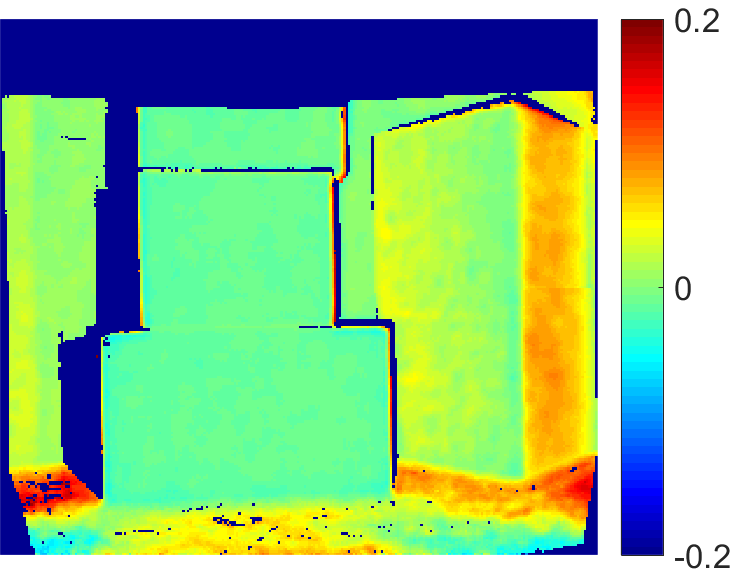}&		 \includegraphics[width=3cm,trim=22ex 8ex 1ex 4ex,clip]{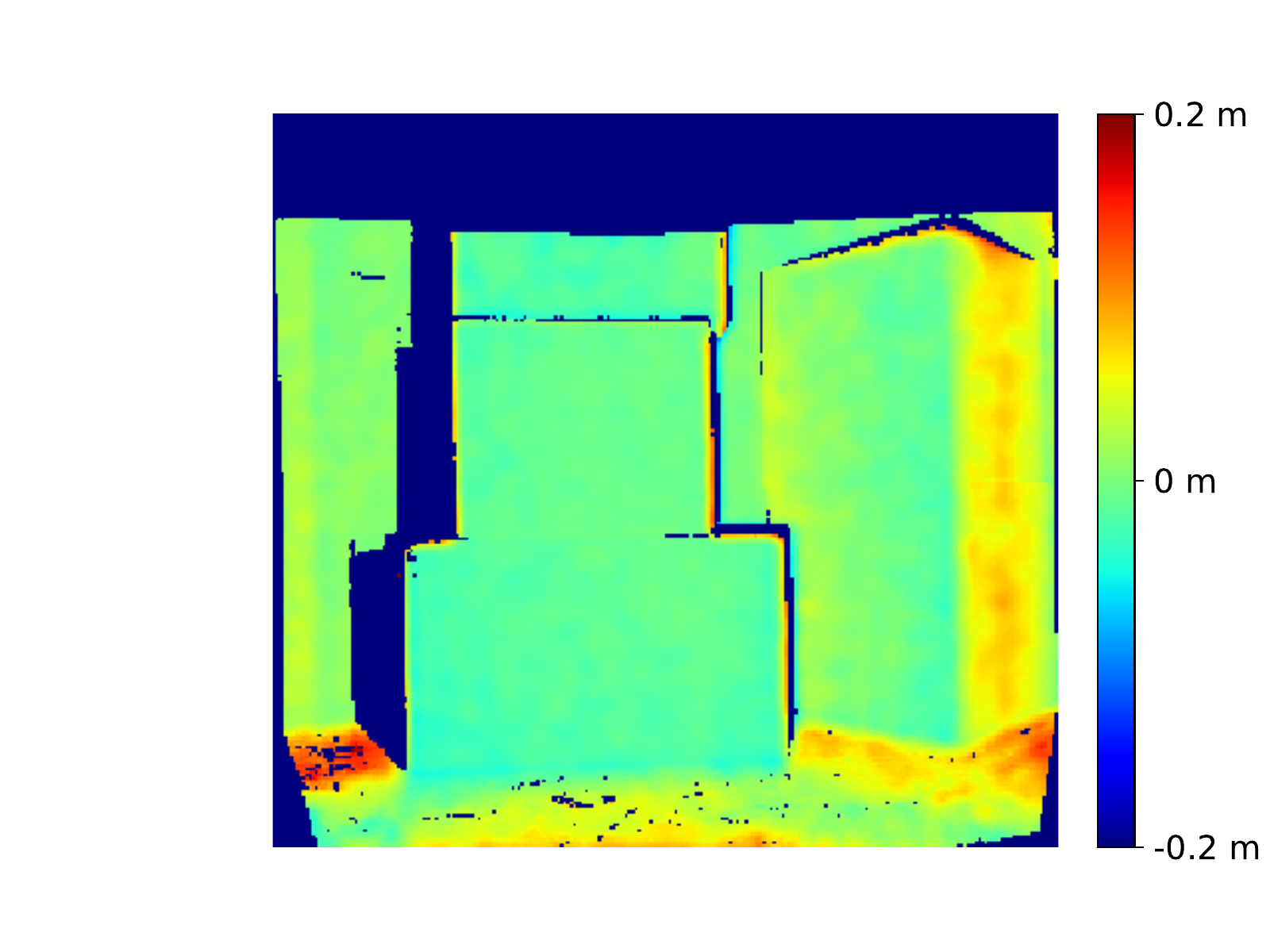}\\	
		\includegraphics[width=2.5cm,trim=21ex 8ex 18ex 11.2ex,clip]{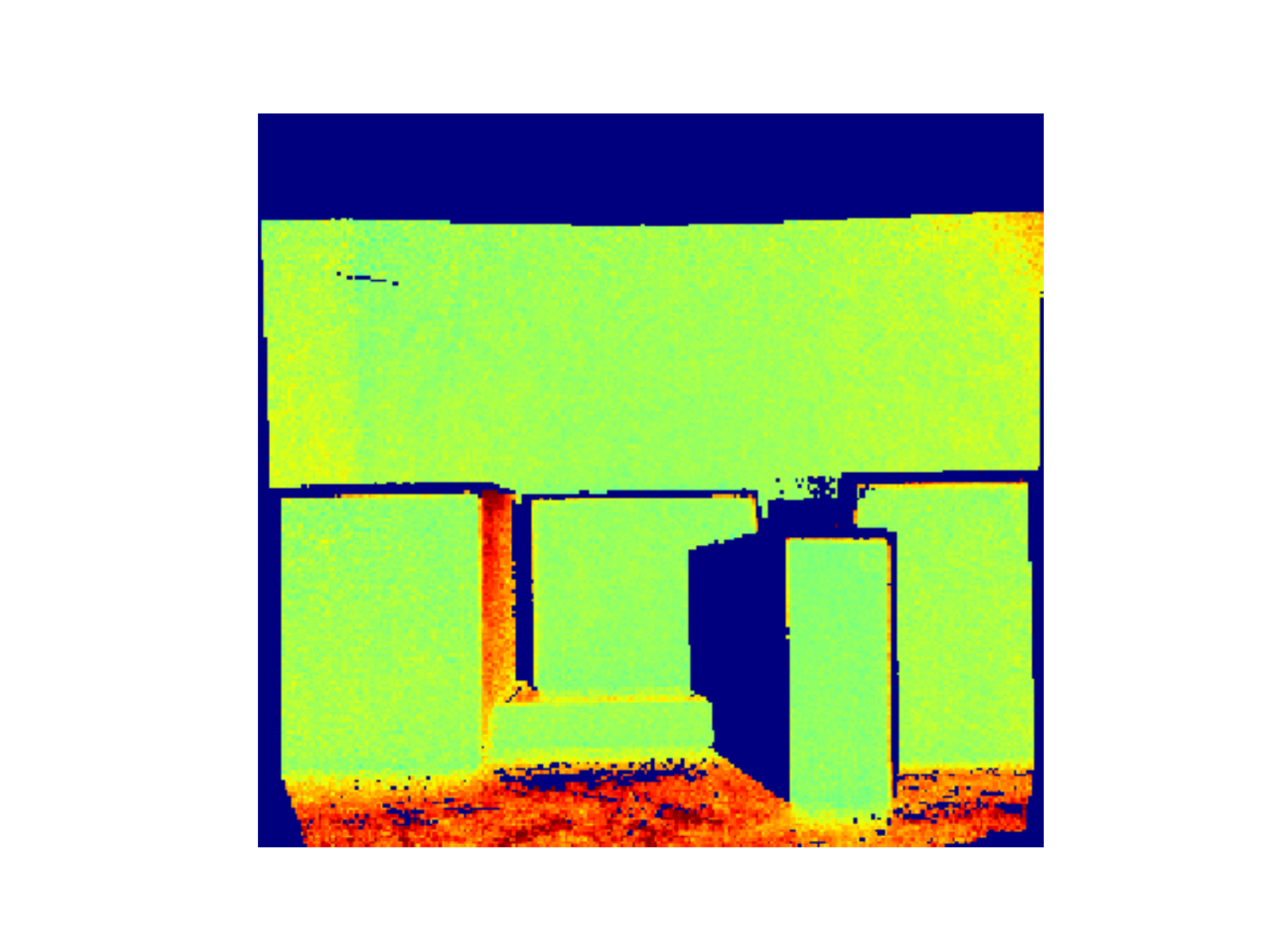}&		 \includegraphics[width=2.5cm,trim=0ex 2ex 15.5ex 2ex,clip]{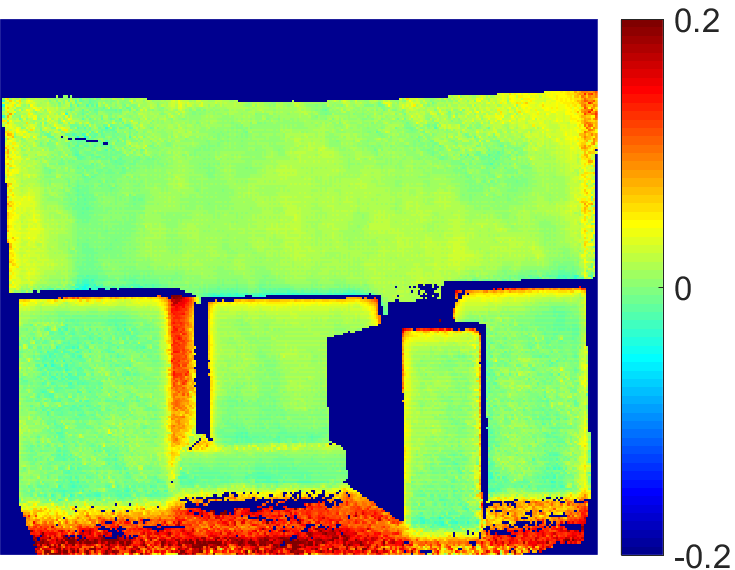}&		 \includegraphics[width=2.5cm,trim=21.7ex 8ex 21ex 7ex,clip]{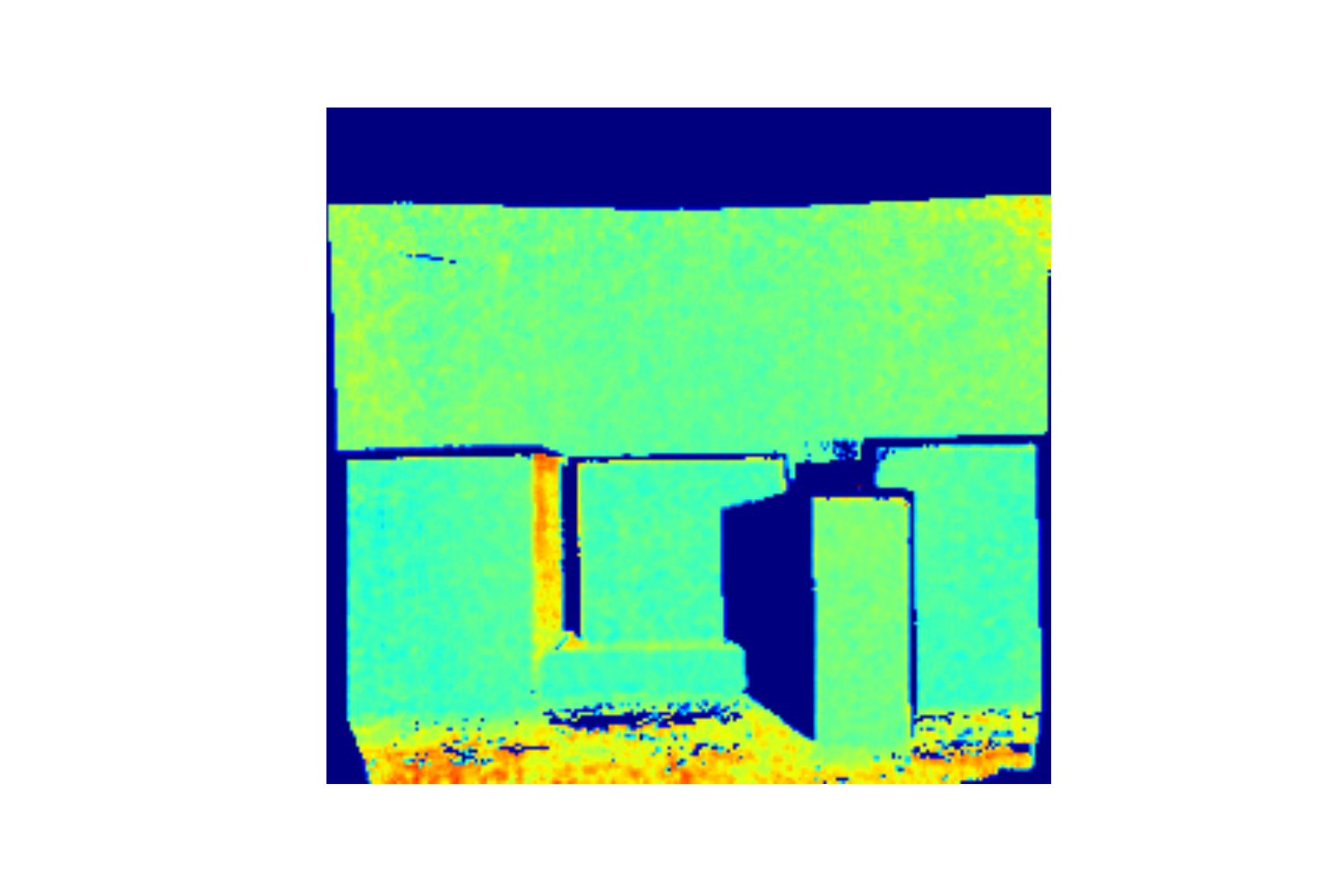}&		 \includegraphics[width=2.5cm,trim=0ex 2ex 15.5ex 2ex,clip]{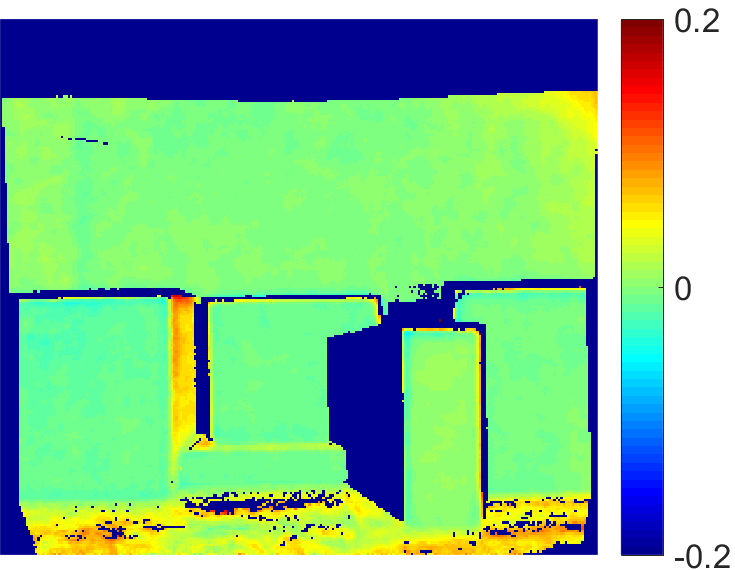}&		 \includegraphics[width=3cm,trim=22ex 8ex 1ex 4ex,clip]{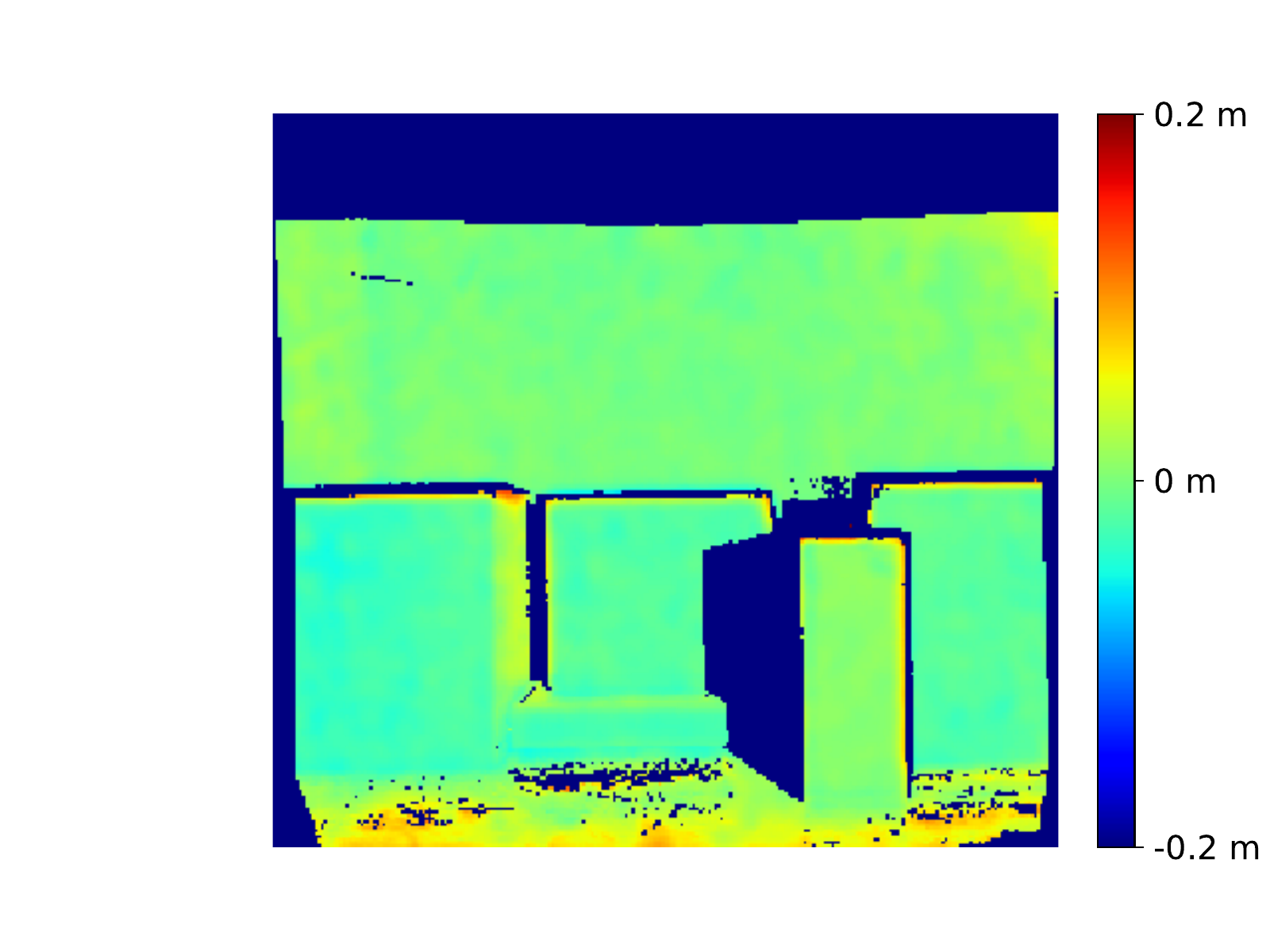}\\	
	\end{tabular}
\caption{Qualitative comparison between some of the best approaches for MPI correction. The first two images come from the $S_4$ dataset, while the other two from $S_5$. All the images display the reconstruction error w.r.t. the ground truth, where green means good reconstruction and red an overestimation.}
\label{fig:comparison} 
\end{figure*}

Figure \ref{fig:comparison} shows a qualitative comparison on a few images from the $S_4$ and $S_5$ datasets. The approach we propose shows a clear improvement on the competitors, providing a good reconstruction also on regions highly corrupted by MPI such as the floor and other steeply sloped scene elements. As an example we can consider the first row of Figure \ref{fig:comparison}, where not only the floor shows a better reconstruction, but the MPI artefacts on the wall on the right are almost completely corrected. Similar considerations can be made on the last image row, where our approach is the only one able to clear the right face of the box on the left side, a particularly difficult surface due to its tilt.

In Table \ref{tab:resultsS1} we report instead the comparison made on the $S_1$ dataset. This is interesting due to the abundant presence of shot noise which can hinder the performance of some approaches. The problem is that, while MPI correction can be performed using only information along the transient dimension, that is not possible for shot noise removal. Networks such as that of Buratto et al. \cite{buratto} and our \textit{D} model have a receptive field of size $3\times3$, which hampers the performance on $S_1$. In this case, as shown in the table, the addition of the \textit{S} module was crucial, as the gap between \textit{SD} and \textit{D} is much wider than before. At the same time however, we are still able to outperform approaches relying on much more complex deep networks, and in particular, the network from Agresti et al. \cite{ag_2018} (that is the work introducing the dataset $S_1$), by more than $1$ cm.

\begin{table}[t]
	\centering
	\small
	\begin{tabular}{lccc}
		\toprule
		Approach   &  $S_4$  & $S_5$  &$\#$ of \\
				& $[cm]$		 & $[cm]$  & parameters\\ 
		\midrule
		Single freq. (60 MHz)   & $5.43$   & $3.62$ & - \\[0.1cm]
		SRA \cite{multi1}		        & $5.11$    & $3.37$ & - \\[0.1cm]
		
		DeepToF \cite{deeptof}	     & $5.13$   & $6.68$ & $330$k \\
		+ calibration						    & $5.46$   & $3.36$  & $330$k\\[0.1cm]
		
		Agresti et al. \cite{ag_2018}   & $3.19$    & $2.22$ &  $150$k\\
		+in-DA \cite{ag_pami}  & $2.40$   & $1.74$  & $150$k\\
		+feat-DA \cite{ag_pami}   & $2.37$   & $1.66$  & $150$k\\
		+output-DA \cite{ag_pami}  & $2.31$   & $\mathbf{1.64}$  & $150$k\\[0.1cm]
		
		Buratto et al \cite{buratto}  & $2.60$   &  $2.12$ & $22$k\\ [0.1cm]
		Ours: \textit{D} (\textit{Walls})   &  $2.46$  & $1.98$ &  $\mathbf{3}$\textbf{k}\\
        Ours: \textit{SD} (\textit{Walls}+$S_1$)   & $\mathbf{2.06}$ & $1.87$ & $23$k \\
		\bottomrule
	\end{tabular}
	\caption{Quantitative comparison between several state-of-the-art MPI correction algorithms on the real datasets $S_4$ and $S_5$. The evaluation metric is the MAE. The complexity of each method is also displayed.}
	\label{tab:results}
\end{table}
\begin{table}[t]
	\centering
	\small
	\begin{tabular}{lcc}
		\toprule
		Approach & $S_1$   &$\#$ of \\
			 & $[cm]$  & parameters\\ 
		\midrule
		Single freq. (60 MHz)	&	$16.7$   & - \\[0.1cm]
		SRA \cite{multi1}	&	 $15.0$       &  - \\[0.1cm]
		
		DeepToF \cite{deeptof} + calibration  &	$26.1$   &  $330$k\\[0.1cm]
		
		Agresti et al. \cite{ag_2018} & $7.49$  &  $150$k\\

		Ours: \textit{D} &  $12.2$  &    $\mathbf{3}$\textbf{k}\\
		Ours: \textit{SD}   & $\mathbf{6.17}$ &  $23$k \\
		\bottomrule
	\end{tabular}
	\caption{Quantitative comparison between several state-of-the-art MPI correction algorithms on the synthetic dataset $S_1$. The evaluation metric is the MAE. The complexity of each method is also displayed.}
	\label{tab:resultsS1}
\end{table}

To conclude, we will now see how our model fares in the presence of extremely high amounts of shot noise. To this aim, we trained our approach on the iToF2dToF dataset \cite{itof2dtof}, which, as described in Section \ref{sec:datasets}, has a very high amount of zero-mean errors. To put things into perspective, the single frequency reconstruction at $100$ MHz of the test set from the measurements with shot noise leads to a MAE of $7.24$ cm, while the same computation done on images with only MPI, gives an error of $1.45$ cm, meaning that MPI accounts only for $20\%$ of the total reconstruction error. 
\begin{figure}
	\centering
  \begin{tabular}{cc}
		1 frequency (100 MHz) & Network prediction\\
		\includegraphics[width=4cm]{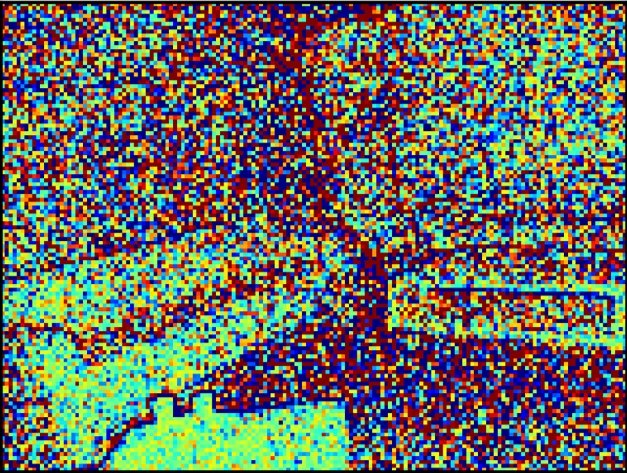}&		 \includegraphics[width=4cm]{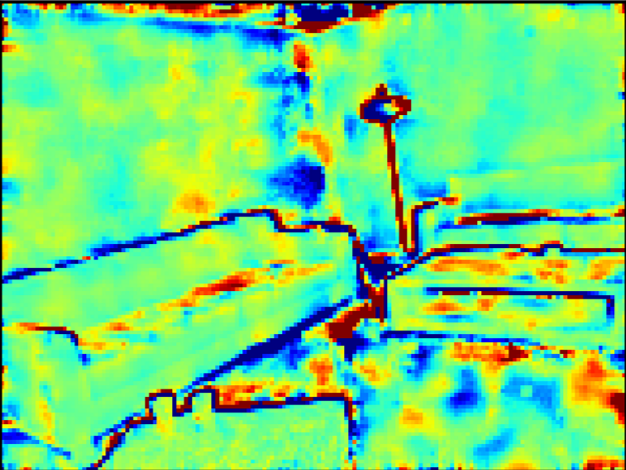}\\
		MAE: 17.53 cm       & MAE: 3.94 cm		 

	\end{tabular}
	
	\caption{Qualitative results on a particularly noisy image from the iToF2dToF dataset \cite{itof2dtof}. On the left side the single frequency reconstruction at 100 MHz, on the right side the network prediction.}
	\label{fig:itof2dtof} 
\end{figure}
Following the setup from \cite{itof2dtof}, we used two input frequencies, $20$ and $100$ MHz, masked the edges using a Canny edge detector during testing and did not consider the highest $1\%$ of errors for the final computation. Our approach shows some remarkable denoising capabilities even in this scenario as it can be seen in Figure \ref{fig:itof2dtof}. Quantitatively, our approach reaches a test error of $1.97$ cm, removing around $75\%$ of the noise, that is behind the performances of iToF2dToF, which removes around $82\%$ of the noise,
but this is to be expected considering the very different scenario. Our approach consists of a very light architecture with extremely good MPI denoising capabilities, which is also able to deal with relatively high amount of shot noise, but the removal of zero-mean noise sources is not its primary objective, while iToF2dToF mostly focuses on this task. Moreover, the difference in complexity between the two architectures is striking: iToF2dToF needs almost two million parameters, while our network is still able to remove three quarters of the total noise using $100$ times less parameters.

\subsection{Transient Reconstruction}
We will now provide some qualitative results on the performance of the \textit{Transient Reconstruction Module}, highlighting its pros and current limitations, and then make a qualitative comparison with iToF2dToF \cite{itof2dtof}, the only other data-driven model that tries to reconstruct transient information. 
\begin{figure}
	\centering
	\begin{tabular}{cc}

		\includegraphics[width=4cm]{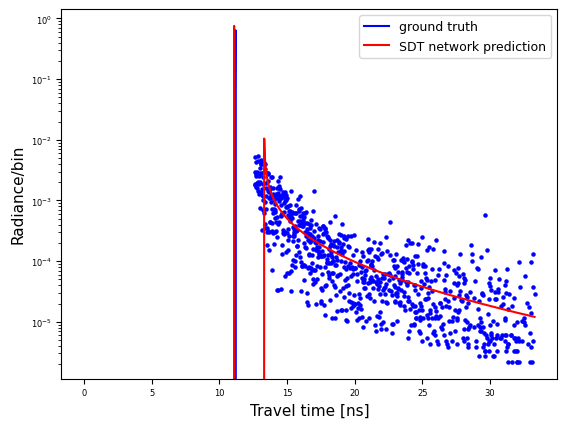}&		 \includegraphics[width=4cm]{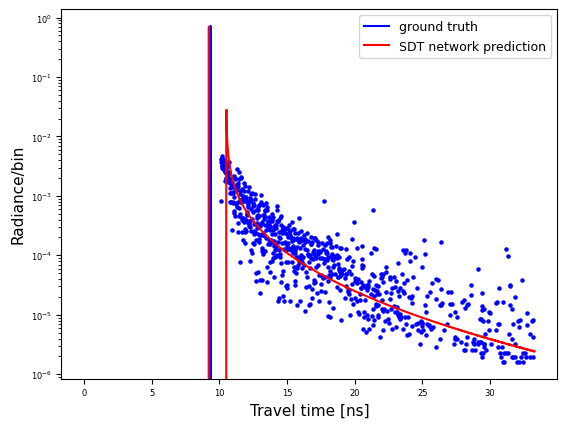}\\
		\includegraphics[width=4cm]{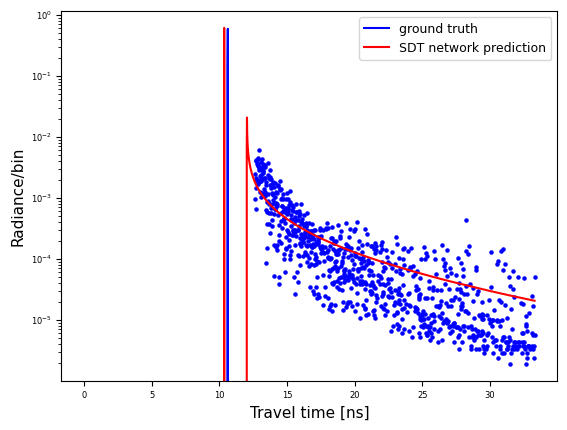}&		 \includegraphics[width=4cm]{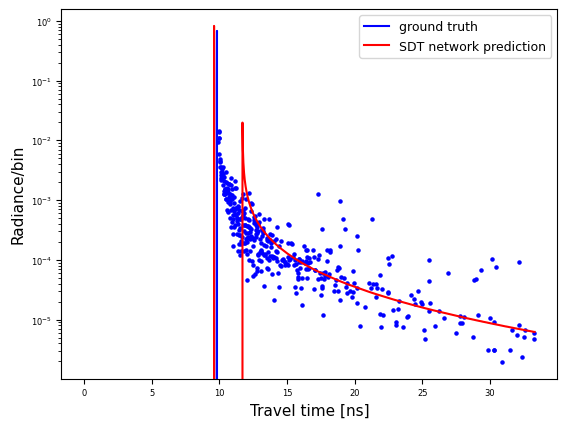}

	\end{tabular}
	
	\caption{Qualitative examples showing the transient reconstruction capabilities of our approach. On the top row we show a pair of good examples, while on the bottom one a pair of less accurate ones. All the plots have a logarithmic scaling.}
	\label{fig:transient} 
\end{figure}
In Figure \ref{fig:transient} we can see a comparison between the transient ground truth and the reconstruction of our network for $4$ pixels from our transient dataset. On the top row we show a pair of good examples, where both the direct and the global components are captured quite well; on the bottom row instead we can see some of the limitations of our model. The direct component still shows a good reconstruction, while the global is more challenging to be reconstructed. The y axis has been logarithmically scaled to show together both the direct and global components.
\begin{figure}
	\centering
	\begin{tabular}{cc}
		
		\includegraphics[width=4cm,trim=4.5ex 0ex 9.5ex 8ex,clip]{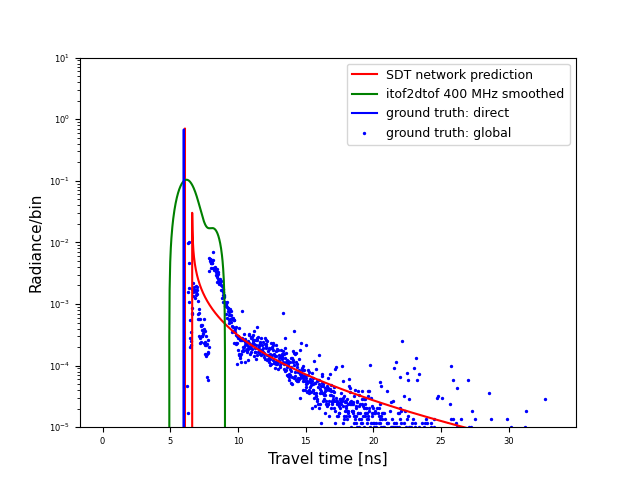}&		 \includegraphics[width=4cm,trim=4.5ex 0ex 9.5ex 8ex,clip]{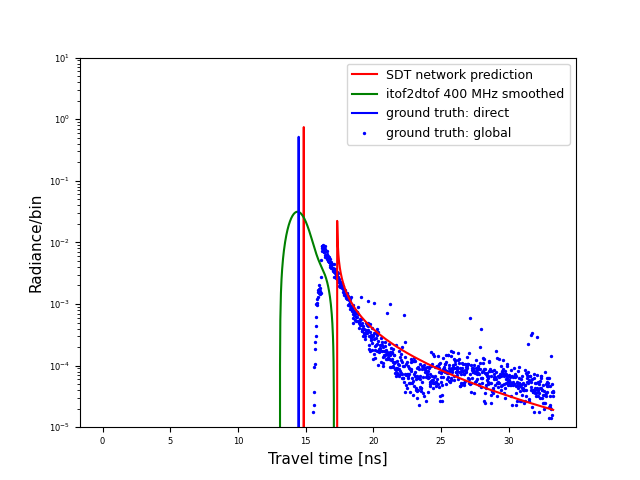}\\

	\end{tabular}
	
	\caption{Qualitative comparison on two pixels from the iToF2dToF dataset.  The direct ground truth has been substituted by a peak whose magnitude consists of the sum of the whole direct, and its position of the weighted average of the direct elements. }
	\label{fig:transient_i2d} 
\end{figure}

In Figure \ref{fig:transient_i2d} we show the performance of our approach on two pixels from the iToF2dToF dataset. Since the direct component of the transient pixel is very spread in this case, and our method only predicts a single peak, we show also an edited version of the ground truth for a better comparison. We substituted the original direct component with a single peak whose magnitude corresponds to the sum of all elements of the original direct, and whose position is taken from the weighted sum of all direct elements' positions, with each weight consisting of the element value itself. We can see that our method shows promising performances also on a previously unseen dataset, capturing very precisely the direct component, and reconstructing reasonably well the global.
It is also clear that our model proposes a much more convincing reconstruction w.r.t. that of iToF2dToF for both components, as the competitor has a much worse estimate, especially for the global component.
\subsection{Ablation studies}
As follows, we show some ablation studies which explain the choices behind our network architecture. Among other variations, we study how the training datasets, the receptive field and the number of input frequencies influence the final performance. 
\paragraph{Training dataset and number of frequencies}
The prediction quality of any data-driven technique heavily depends on the goodness of the dataset used to train it in the first place. For this reason, we decided to test three different scenarios: in the first one we trained our \textit{SD} model on the $S_1$ dataset alone, in the second one the \textit{Walls} dataset was the only input of the network, and in the final one we used both for supervision, as we did for the results in the previous section.  In order to make the comparison fair, we also decided to perform different trainings on the \textit{Walls} dataset supervising either on $\boldsymbol{v_d}$ or on $\boldsymbol{\varphi_d}$ (i.e., the phase of the direct component). Finally, as the two datasets have a different size, we also added one entry where our method has been trained on a reduced version of our dataset ($45$ training images instead of $134$). The outcome of this study is shown in Table \ref{tab:ablation_dataset}.
\begin{table}[h]
	\centering
	\small
	\begin{tabular}{lccccc}
		\toprule
		Training  &  \multicolumn{2}{c}{$\#$ of images} & $S_1$  &  $S_4$  & $S_5$ \\
		 dataset  & $S_1$ & \textit{Walls} & $[cm]$		& $[cm]$		 & $[cm]$  \\ 
		\midrule
		$S_1$ ($\boldsymbol{\varphi_d}$)	& $40$& -   &	$\mathbf{5.93}$   & $3.65$   & $2.33$ \\
		\textit{Walls}	($\boldsymbol{\varphi_d}$) &	 -& $45$    &    $19.2$     &  $2.46$   &  $2.39$   \\ 				
		\textit{Walls}	($\boldsymbol{\varphi_d}$)	&- & $134$     &  $18.4$       & $2.34$    & $2.40$ \\ 		
		\textit{Walls}	($\boldsymbol{v_d}$)	  &- & $134$  &  $12.2$       & $2.26$    & $2.44$   \\ 
		
		$S_1$ ($\boldsymbol{\varphi_d}$) + \textit{Walls} ($\boldsymbol{\varphi_d}$)	 &  $40$ & $134$      &	$6.99$    & $2.26$   & $2.04$ \\		
		$S_1$ ($\boldsymbol{\varphi_d}$) + \textit{Walls} ($\boldsymbol{v_d}$)	&  $40$ & $134$     &	$6.17$    & $\mathbf{2.06}$   & $\mathbf{1.87}$    \\

		\bottomrule
	\end{tabular}
	\caption{Quantitative comparison of the performance of different training datasets on the synthetic dataset $S_1$ and on the real datasets $S_4$ and $S_5$ for different training datasets and losses.  The evaluation metric is the MAE. All trainings have been performed on the \textit{SD} network.}
	\label{tab:ablation_dataset}
\end{table}
Considering first the results on $S_1$, we can see that the training performed on $S_1$ itself leads to the best performance, followed at a short distance by using both datasets; training on \textit{Walls} alone instead falls behind by a large margin. This is not surprising as the images from \textit{Walls} have no shot noise, thus explaining the poor performance. What's more remarkable instead is the prediction on real data, as the $S_1$ dataset yields a significantly poorer performance when compared to the training on \textit{Walls}. The dissimilarity between the two datasets is striking: the former is made of much more complex scenes, shows a wide range of textures and has a good amount of simulated noise; the latter instead focuses on extremely simple structures, no changes in texture and its only noise source is MPI.
What seems to be happening is that the added complexity of the dataset and some issues it presents (some scene elements of the $S_1$ dataset present very unreliable information), make the prediction harder for the network. Moreover, our approach relies mostly on the information in the transient direction, making in this way the complex structures from $S_1$ less relevant than the cleaner phasor data from \textit{Walls}. In the end, combining the two datasets leads to the best overall solution, with a competitive performance on $S_1$, and the best prediction on both real datasets. It is also useful to point out that using $\boldsymbol{v_d}$ for supervision improves on a training only based on the phase, showing how a transient dataset can be particularly useful for MPI correction.

Another point of interest concerns the ability of the model in dealing with a different number of input frequencies. In particular, we decided to train our model with two frequencies, $20$ and $50$ MHz, and see how it coped in comparison to the three frequencies input. In Table \ref{tab:ablation_frequencies} we can see that the lack of the $60$ MHz component (depth estimated from higher frequencies has typically a smaller error) has indeed a toll on the overall performance, but the model is still able to clean a noteworthy amount of MPI. To put things into perspective, it is enough to look at Table \ref{tab:ablation_dataset}, where we can see how a $2$ frequencies training on \textit{Walls} still provides a better prediction than a $3$ frequencies training on $S_1$ on the real datasets.
\begin{table}[h]
	\centering
	\small
	\begin{tabular}{lccc}
		\toprule
		Input frequencies & $S_1$ dataset &  $S_4$ dataset & $S_5$ dataset\\
		& $[cm]$		& $[cm]$		 & $[cm]$  \\ 
		\midrule
		Single frequency $50$ MHz  &   $18.0$    &  $5.31$   &  $3.82$  \\
		$20,50$ MHz	&	$6.56$   & $3.44$   & $2.31$  \\
		$20,50,60$ MHz	     &	$\mathbf{6.17}$    & $\mathbf{2.06}$   & $\mathbf{1.87}$  \\
		
		\bottomrule
	\end{tabular}
	\caption{Quantitative comparison of the performance of a different number training frequencies on the synthetic dataset $S_1$ and on the real datasets $S_4$ and $S_5$. The evaluation metric is the MAE. The first row shows the baseline error at $50$ MHz without any processing.} 
	\label{tab:ablation_frequencies}
\end{table}
\paragraph{Receptive field and network complexity}
As we have seen in Table \ref{tab:results}, there is no clear correlation between the number of parameters of the architecture and its actual performance. This is due to multiple factors, such as the high risk of overfitting due to the domain difference between training and test data, the relatively small sizes of the datasets (even the \textit{Walls} one only comprises around $200$ images) and the main focus of the model itself, which can be centered on the use of spatial features (e.g. \cite{ag_2018,ag_unsup}) or on the transient dimension (\cite{buratto} and ours). In our case, we have to consider an additional factor which are the characteristics of our training datasets. In particular, while  a relatively large receptive field would be better in order to deal with shot noise, we cannot enlarge it too much as our main tool, the \textit{Walls} dataset, is made mostly of flat surfaces and there is a risk of overfitting its structure. Learning too much from this dataset geometry, as it can be seen in Table \ref{tab:ablation_receptive}, decreases the performance of the model. From the Table we can see that the overall best performance on the datasets arises from a receptive field of either $11\times11$ or $15\times15$, while it clearly degrades for smaller or bigger sizes. We decided to employ a receptive field of $11\times11$ due to its slightly better performance and the reduced network complexity.
\begin{table}[h]
	\centering
	\small
	\begin{tabular}{cccc}
		\toprule
		Receptive field & $S_1$ dataset &  $S_4$ dataset & $S_5$ dataset\\
		& $[cm]$		& $[cm]$		 & $[cm]$  \\ 
		\midrule

		$7\times7$	     &	 $8.08$  &  $2.44$ & $\mathbf{1.82}$\\	
		$11\times11$	     &	$\mathbf{6.17}$    & $2.06$   & $1.87$  \\		
		$15\times15$	     &	 $6.35$   &  $\mathbf{2.00}$  & $2.10$  \\		
		$21\times21$	     &	 $8.19$   &  $2.42$  &  $2.45$ \\
		
		\bottomrule
	\end{tabular}
	\caption{Quantitative comparison of the performance of different receptive fields for the \textit{SD} network on the synthetic dataset $S_1$ and on the real datasets $S_4$ and $S_5$. The evaluation metric is the MAE.}
	\label{tab:ablation_receptive}
\end{table}



%% file: sections/conclusions.tex
\section{Conclusions}
\label{sec:conclusions}
In this work we have introduced a novel network for MPI denoising and transient reconstruction. The architecture is modular: the \textit{Spatial Feature Extractor} is useful to deal with zero mean errors, the \textit{Direct Phasor Estimator} deals with MPI and the \textit{Transient Reconstruction Module} reconstructs the transient information from the previous. We have proposed two very compact networks, and compared their performance against some of the best models in the literature. Our \textit{SD} architecture reaches state-of-the-art performance both on synthetic and real data, while the \textit{D} one shows comparable performance, while only needing $3$k network weights. The model shows also promising results regarding the reconstruction of transient information, but still has a few limitations that we plan to address in future works. 
A key challenge that we will explore is how to find an accurate  model for the global component that can be represented with a few parameters.
Moreover, we plan to substitute the parametric functions with a  network in order to learn more complex global component shapes.